%% file: GEL_paper.tex
\documentclass[10pt,twocolumn,letterpaper]{article}

\usepackage[pagenumbers]{cvpr} %
\usepackage[utf8]{inputenc} %
\newcommand{\fid}{\mbox{Fr\'{e}chet Inception Distance }}
\newcommand{\od}{{v}}

\newcommand{\chuck}{s} %
\newcommand{\keep}{y} %
\usepackage{graphicx}
\usepackage{amsmath}
\usepackage{amssymb}
\usepackage{booktabs}
\input{math_commands}

\usepackage{mathtools}
\DeclarePairedDelimiterX{\infdivx}[2]{(}{)}{%
  #1\;\delimsize\|\;#2%
}
\newcommand{\kl}{D\infdivx}
\newcommand{\klkl}{D_{KL}\infdivx}
\usepackage{amsfonts}       %
\usepackage{nicefrac}       %
\usepackage{microtype}      %
\usepackage{xcolor}         %

\input{figures/least_likely_samples_commands}
\usepackage[frozencache=true,cachedir=mintedoutput]{minted}
\usepackage{algorithm}
\usepackage{algorithmic}
\usepackage{multicol}
\usepackage{adjustbox}
\usepackage{amsthm}
\usepackage{thmtools, thm-restate}
\usepackage{multirow}
\newcommand{\suman}[1]{\textcolor{red}{\textit{[S: #1]}}}
\renewcommand{\suman}[1]{\textcolor{red}{#1}}
\renewcommand{\suman}[1]{#1}
\newcommand{\KGEL}{\textcolor{blue}{KGEL}}
\renewcommand{\KGEL}{KGEL}
\newcommand{\ktsgel}{\textcolor{blue}{KGEL2}}
\renewcommand{\ktsgel}{KGEL2}
\newcommand{\melanie}[1]{}
\newcommand{\remi}[1]{}
\newcommand{\marc}[1]{}
\newcommand{\X}{\textcolor{red}{x}}
\renewcommand{\X}{x}
\newcommand{\Y}{\textcolor{red}{y}}
\renewcommand{\Y}{y}
\declaretheorem{theorem}

\usepackage[pagebackref,breaklinks,colorlinks]{hyperref}
\usepackage{url}
\usepackage[accsupp]{axessibility}

\usepackage[capitalize]{cleveref}
\crefname{section}{Sec.}{Secs.}
\Crefname{section}{Section}{Sections}
\Crefname{table}{Table}{Tables}
\crefname{table}{Tab.}{Tabs.}

\newcommand{\citep}{\cite}
\newcommand{\citet}{\cite}

\theoremstyle{plain}

\theoremstyle{definition}

\theoremstyle{remark}

\def\cvprPaperID{5093} %
\def\confName{CVPR}
\def\confYear{2023}

\begin{document}

\title{Understanding Deep Generative Models with Generalized Empirical Likelihoods}

\author{Suman Ravuri, M\'elanie Rey, Shakir Mohamed\\
DeepMind\\
London, UK\\
{\tt\small \{ravuris,melanierey,shakir\}@deepmind.com}
\and
Marc Peter Deisenroth\\
University College London\\
London, UK\\
{\tt\small m.deisenroth@ucl.ac.uk}
}
\maketitle
\input{00_abstract}
\input{01_intro_cvpr}

\input{02_moments_and_gel}
\input{03_gen_models_tests}
\input{04_experiments}
\input{05_related_work}
\input{06_discussion}

{\small
\bibliographystyle{ieee_fullname}
\bibliography{el_refs}
}
\appendix
\input{07_appendix}
\end{document}

%% file: math_commands.tex
\usepackage{amsmath,amsfonts,bm}

\def\eqref#1{equation~\ref{#1}}

\def\1{\bm{1}}

\def\vc{{\bm{c}}}

\def\vm{{\bm{m}}}

\def\vx{{\bm{x}}}
\def\vy{{\bm{y}}}
\def\vz{{\bm{z}}}

\DeclareMathAlphabet{\mathsfit}{\encodingdefault}{\sfdefault}{m}{sl}
\SetMathAlphabet{\mathsfit}{bold}{\encodingdefault}{\sfdefault}{bx}{n}

\newcommand{\E}{\mathbb{E}}

%% file: 00_abstract.tex
Understanding how well a deep generative model captures a distribution of high-dimensional data remains an important open challenge. It is especially difficult for certain model classes, such as Generative Adversarial Networks and Diffusion Models, whose models do not admit exact likelihoods. In this work, we demonstrate that generalized empirical likelihood (GEL) methods offer a family of diagnostic tools that can identify many \suman{deficiencies} of deep generative models (DGMs). We show, with appropriate specification of moment conditions, that the proposed method can identify \emph{which} modes have been dropped, the degree to which DGMs are mode imbalanced, and whether DGMs sufficiently capture intra-class diversity. We show how to combine techniques from Maximum Mean Discrepancy and Generalized Empirical Likelihood to create not only distribution tests that retain per-sample interpretability, but also metrics that include label information. We find that such tests predict the \suman{degree} of mode dropping and mode imbalance up to $60\%$ better than metrics such as improved precision/recall. We provide an implementation at \url{https://github.com/deepmind/understanding_deep_generative_models_with_generalized_empirical_likelihood/}.

%% file: 01_intro_cvpr.tex
\section{Introduction}
\looseness=-1 In an era that has witnessed deep generative models (DGMs) produce photorealistic images from text descriptions \citep{dalle2, nokey, saharia2022photorealistic}, speech rivaling that of professional voice actors \citep{oord2016wavenet, child2019generating, goel22a}, and text seemingly indistinguishable from writing on the internet \citep{brown2020language, openai2023gpt4}, it perhaps seems quaint to focus on better evaluation of such models. One might reasonably claim that such evaluation may have been useful prior to the development of these models, but such evaluation is certainly less important now. They might further bolster their claim by noting that researchers have already proposed reasonable metrics for these modalities --- \fid for images \citep{heusel2017gans}, Mean Opinion Score (MOS) for speech, and \suman{perplexity} for text ---, and the results have been so compelling that researchers and practitioners have begun to deploy such models in downstream tasks \citep{noe2019boltzmann, boyda2021sampling, wirnsberger2021normalizing, albergo2021flow}. So why should we focus on better evaluation now?
\input{figures/least_likely_sample_single_figure/class_43_figure}

We are advocating for better and more nuanced evaluation now precisely \emph{because} DGMs have reached sufficient maturity to be used in downstream tasks. When Deep Generative Models struggled to produce realistic $32\times32$ images, a researcher working on the model may not have found nuanced evaluation useful. Now that DGMs can credibly produce realistic \suman{megapixel} images from text input, that same researcher may find more nuanced evaluation helpful to understand what deficiencies still exist. %
This view reflects recent trends in generative model evaluation, as researchers have now started to adopt metrics such as precision/recall \citep{kynkaanniemi2019improved} to better understand \emph{how} the model is misspecified. %

\looseness=-1 Our approach comprises three parts: a moment condition on the DGM and data distributions that tells us if the model captured a salient property of the data distribution; a score that indicates how well a moment condition is satisfied; and a decomposition of that score that shows how each test point contributed. %
From these three elements, we can diagnose how a DGM is misspecified.

We observe that many %
evaluation metrics can be expressed as a set of moment conditions --- for example, \fid can be recast as the condition that the first and second moments of Inception v3 Pool3 (further denoted as ``Pool3'') features match. Then, with a set of moment conditions as our specification, we employ the machinery of Empirical Likelihood methods \citep{owen1990empirical} to provide us both the aggregate and per-sample scores.

The Empirical Likelihood is a moment condition test that approaches the problem by answering the following question: how much must the data distribution change in order to satisfy the moment condition? Typically, researchers use these tests to determine whether a moment condition is satisfied. %
We find that beyond just identifying \emph{whether} a moment condition is satisfied, under certain conditions on the moments, %
the changed distribution can tell us \emph{how} a DGM is misspecified. Our main contribution is elucidating the conditions under which we can use Empirical Likelihood methods as a diagnostic tool. %
In the process, we introduce four new tests and show the following:

\textbf{Mode dropping.} We show how to create moment conditions that identify \emph{which} modes have dropped, without access to label information from the model.

\looseness=-1 \textbf{Mode imbalance.} \suman{We experimentally show that when a DGM samples one mode more frequently than others, GEL methods can predict the degree of this mode imbalance.}

\textbf{Interpretable tests.} We detail how to use moment conditions from Maximum Mean Discrepancy \citep{gretton2012kernel} approaches to create %
interpretable tests.

\suman{\textbf{Improper label conditioning.} By including label information, we create tests that identify when a label-conditioned generative model ignores its conditioning label.}

%% file: figures/least_likely_sample_single_figure/class_43_figure.tex
\begin{figure}[t]
    \centering
    \includegraphics[width=\linewidth]{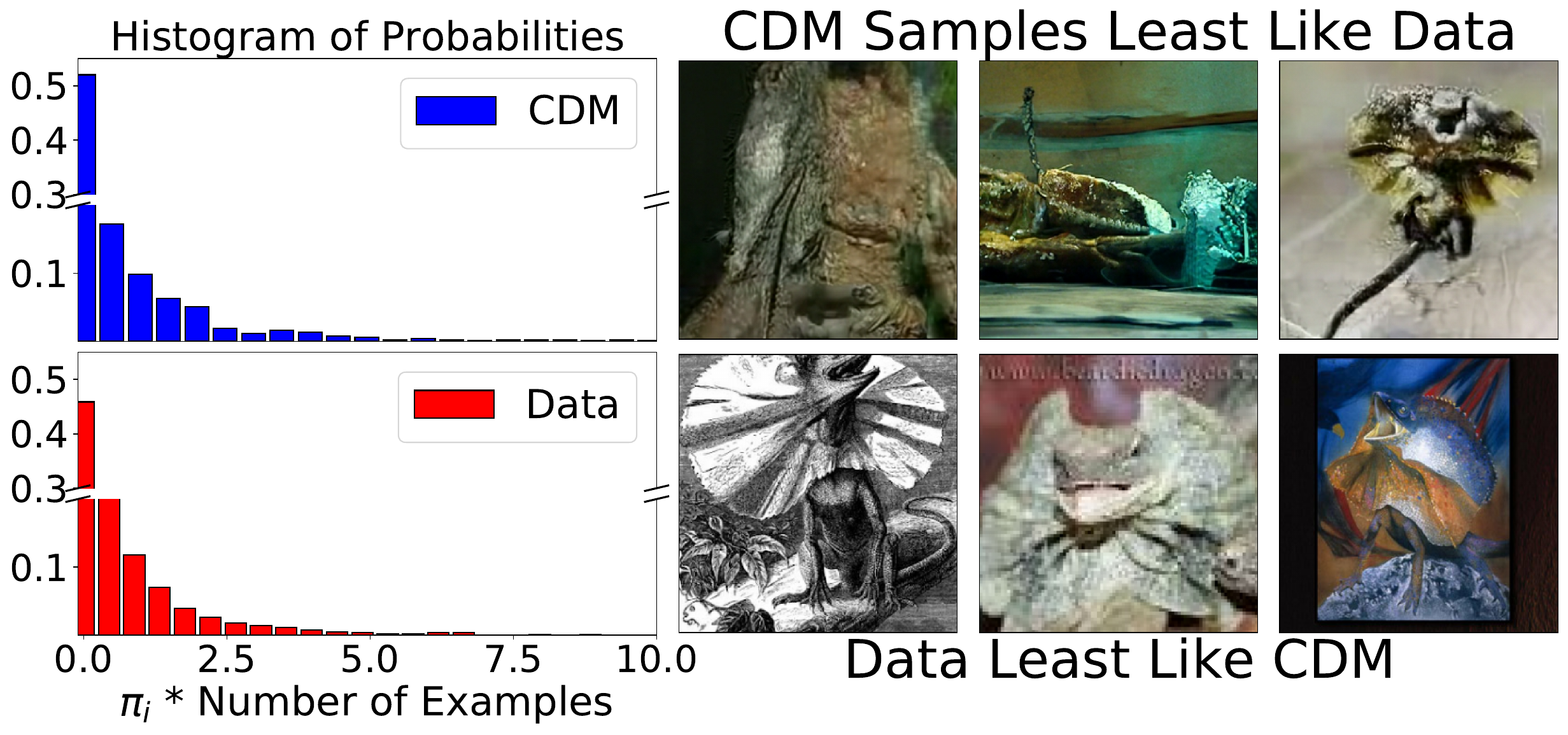}
    \caption{Empirical Likelihood methods can identify model samples outside of the data distribution, and data samples outside of the model distribution. In this example, we use a two-sample generalized empirical likelihood test to evaluate how well a Cascaded Diffusion Model trained on ImageNet captures the ``Frilled Lizard'' class. The three samples on the bottom right show samples from the data distribution the model is not likely to represent, and those on the top right show samples from the model not likely to be in the data distribution. These examples have 0 probability in the empirical likelihood test.}
    \vspace{-5mm}
    \label{fig:least_likely_class_210}
\end{figure}

%% file: 02_moments_and_gel.tex
\section{Background}
\label{sec:gel_intro}
In this section, we provide a brief background on the Empirical Likelihood and moment conditions useful for generative model evaluation. %
Both of these areas have a rich history of their own. We refer the reader to the excellent monograph by \citet{owen2001empirical} for a broader discussion on Empirical Likelihood and \cite{alma991043249379903276} for moment restrictions. 
\input{figures/background_tables/background_table}

\looseness=-1\textbf{Generalized Empirical Likelihood} The Empirical Likelihood (EL) \citep{owen1990empirical} is a classical nonparametric method of statistical inference. Originally, EL was proposed as a method of inference on the mean: given $n$ independent samples $x_1, \dots, x_n \in \mathbb{R}^d$ from unknown distribution $p$, EL determines whether the mean of $p$ is equal to a known constant $\mathbf{c} \in \mathbb{R}^d$.
EL is unique in its approach, as it models the samples with a weighted empirical distribution $P_{\boldsymbol{\pi}}(x) = \sum_{i=1}^n \pi_i \mathbb{I}_{[x_i=x]}$, with weights $\pi_i$ that satisfy the rules of probability (namely that $\sum_i \pi_i = 1, \pi_i \geq 0$). It finds, among all distributions $P_{\boldsymbol{\pi}}$ that match the mean condition $\E_{P_{\boldsymbol{\pi}}}[\X] \equiv \sum_{i=1}^n \pi_i x_i = \vc$, the one that maximizes the likelihood $\prod_i \pi_i$. This ``empirical likelihood'' can be expressed as the solution to the convex problem
\begin{align}
\begin{aligned}
\label{eq:el}
    &\max_{\{\boldsymbol{\pi} | \sum_i \pi_i=1, \pi_i \geq 0\}} \sum_{i=1}^n \log{\pi_i} %
    &\mbox{s.t.}\quad
    \E_{\X \sim P_{\pi}}[\X] = \vc %
    \end{aligned}
\end{align}
\looseness=-1 We denote by $\boldsymbol{\pi}^*$ and $P_{\boldsymbol{\pi}^*}$ the weights and implied distribution, respectively, that solve \cref{eq:el}\suman{, and abbreviate ``\emph{subject to}'' to ``\emph{s.t.}''}.
Intuitively, if $\E_{\X \sim p}[\X]= \vc$, then we expect $P_{\boldsymbol{\pi}^*}$ to be ``close'' to the empirical distribution $\hat{P}_n(x) \equiv \sum_{i=1}^n n^{-1} \mathbb{I}_{[\X_i=x]}$.\footnote{In absence of the mean constraint $\sum_{i=1}^n \pi_i \X_i = \vc$, $\pi^* = n^{-1}\mathbf{1}$, the implied distribution $P_{\boldsymbol{\pi}^*}(x)$ is the empirical distribution, also known as the nonparametric maximum likelihood estimate of the sample \citep{kiefer1956consistency}.} \cref{eq:el} makes this notion precise by measuring the closeness of the two distributions with the KL divergence $\klkl{\hat{P}_n}{P_{\boldsymbol{\pi}}}$. If no distribution satisfies the mean constraint (which is equivalent to the mean not lying in the interior of the convex hull of $\{x_i\}$\footnote{\cref{ssec:convex_hull_condition} includes further discussion of the convex hull condition.}), then by convention the empirical likelihood and KL divergence are $-\infty$ and $\infty$, respectively.

We use two extensions for generative model evaluation. The first extension is replacing the mean condition $\E_{P_{\pi}}[\X] = \vc$ with a moment condition $\E_{\X \sim P_{\pi}}[\vm(\X; \vc)] = \mathbf{0}$ \cite{qin1994empirical}, making the method much more general. %
The second is replacing the KL divergence with one from the Cressie-Read family\footnote{This family of divergences are of the form $CR(\lambda) = \frac{2}{\lambda(\lambda+1)} \sum_i [(n\pi_i)^{-\lambda} - 1]$.} \citep{cressie1984multinomial}. Using other members relaxes the condition that $\pi_i$ be strictly positive, and is useful \suman{for identifying mode dropping}. The resulting objective is called the \emph{Generalized Empirical Likelihood} (GEL): 
\begin{equation}
\label{eq:el_stat_gee}
    \min_{\{\boldsymbol{\pi} | \sum_i \pi_i =1, \pi_i \geq 0\}} \kl{\hat{P}_n}{P_{\pi}} ~~%
    \mbox{s.t.}~~ \E_{\X \sim P_{\pi}}[\vm(\X; \vc)] = \mathbf{0}
\end{equation}
\looseness=-1 In this work, we use two members of the family, shown in the top panel of \cref{tab:EL_background}: the ``exponential tilting'' (ET) objective, and the ``Euclidean likelihood''.\footnote{The latter objective is mainly of intellectual interest: as the solution is proportional to the Hotelling T-square statistic \citep{hotelling1931generalization}, we can recast any T-square statistic as a GEL.} %
We also use a third extension, a two-sample version of the above test, but we defer discussion of this extension to \cref{ssec:proposed_test}.

\looseness=-1 \textbf{Moment Conditions} \suman{We note that many evaluation metrics are statistics of moments.} \suman{They may be as simple as checking if low-order moments match.
Indeed, many popular metrics for generative model evaluation are statistics of a low-order of moments: for example, \fid and Kernel Inception Distance are statistics of the first two and three moments, respectively.} \suman{They may also be as complex as a full distribution test: for a likelihood-based model, the condition that $\E_x[\nabla_{\theta}\log p_{\theta} (x)] = \boldsymbol{0}$ implies the likelihood is maximized.}
\suman{By decoupling the moment condition from the statistic, we can construct alternative tests based on GEL objectives.}
The bottom of \cref{tab:EL_background} shows some common examples and the implied moment conditions.

There exist two issues with applying the Empirical Likelihood to the moment conditions in the table: with the exception of the score function, \suman{the moment conditions are not sufficient to distinguish between two distributions;}
and the dimensionality of the moment conditions may be extremely high, leading to statistical and computational challenges. While the first issue might not be particularly problematic, as we may be more interested in usability \suman{than in} theoretical robustness, the second issue is more \suman{pressing}, as we aim to keep the dimensionality relatively low.

A particularly appealing set of moment restrictions that addresses these issues are those \suman{given by} the Maximum Mean Discrepancy (MMD) \citep{gretton2012kernel}
\begin{equation}
\begin{gathered}
\label{eq:mmd2}
    D^2(p, q) = \!\!\mathop{\E}_{x_1, x_2}\!\![k(x_1, x_2)] +\!\!\! \mathop{\E}_{y_1, y_2}\!\![k(y_1, y_2)] -2\mathop{\E}_{x, y}[k(x, y)]%
\end{gathered}
\end{equation}
where $x, x_1, x_2 \sim p$ and $y, y_1, y_2 \sim q$.
With the appropriate choice of kernel --- such as when the kernel is characteristic \citep{fukumizu2004dimensionality} or universal --- $D^2(p, q) = 0$ iff $p=q$, and, unlike competing approaches, unbiased estimators exist. Despite these appealing properties, metrics based on MMD, such as Kernel Inception Distance (KID) \citep{binkowski2018demystifying}, perhaps due to the $O(n^2)$ computational cost, have not enjoyed widespread adoption. More importantly, a straightforward inclusion of the MMD constraint into \cref{eq:el_stat_gee} is computationally difficult, as the constraint is \suman{a nonlinear function of $\pi_i$}.

\looseness=-1 \suman{Instead, we focus on an alternative characterization based on Mean Embeddings (ME) \citep{chwialkowski2015fast, jitkrittum2016interpretable}. In this approach, one tests equality of distributions $p$ and $q$ by comparing \emph{mean embeddings} $\mu_p(t) \equiv \E_x[k(x, t)]$ and $\mu_q(t) \equiv \E_y[k(y, t)]$, where $t$ is a \emph{witness point} sampled from a third distribution $r$. A somewhat remarkable property is that, if $k$ is characteristic, analytic, and integrable, and $r$ is absolutely continuous with respect to the Lebesgue measure, $\sum_w (\mu_p(t_w) - \mu_q(t_w))^2 > 0$ a.s. if $p\neq q$ and $\sum_w (\mu_p(t_w) - \mu_q(t_w))^2 = 0$ a.s. if $p=q$.}\footnote{\suman{One can interpret this condition colloquially as satisfying the axiom of coincidence with probability 1. The aforementioned work establishes that the statistic also satisfies the other axioms of a distance ``with probability 1'' (this notion of distance is called a \emph{random metric}).}}

\label{ssec:kernel_v1}
\suman{Previous works \citep{chwialkowski2015fast, jitkrittum2016interpretable} define a semimetric $n\bar{\vz}^\top \Sigma_{\vz}^{-1} \bar{\vz}$, where $\bar{\vz} \equiv \frac{1}{n} \sum_i \vz_i$, $\Sigma_{\vz} \equiv \frac{1}{n-1} \sum_i (\vz_i - \bar{\vz})(\vz_i - \bar{\vz})^\top$} and $\vz_i = [k(x_i, t_1) - k(y_i, t_1), \dots, k(x_i, t_W) - k(y_i, t_W)]^\top$.
\suman{This T-square statistic can be written as the solution (up to a constant) to the Euclidean likelihood with moment condition $\E_{\vz}[\vz] = \boldsymbol{0}$.}
\begin{equation}
\label{eq:ME_euc_lik}
\begin{gathered}
\min_{\{\boldsymbol{\pi} | \sum_i \pi_i=1\}} \frac{1}{2} \sum_{i=1}^n \left (\pi_i - n^{-1} \right )^2
\quad \mbox{subject to}~ \\ 
\sum_{i=1}^n \pi_i %
{\scriptscriptstyle \begin{bmatrix}
k(x_i, t_1) - k(y_i, t_1)\\
\vdots \\
k(x_i, t_W) - k(y_i, t_W)
\end{bmatrix}} = \boldsymbol{0} %
\end{gathered}
\end{equation}
\suman{Note that the moment condition is linear in $\boldsymbol{\pi}$ and is within the GEL framework. In the following section, we modify \suman{\cref{eq:ME_euc_lik}} to better diagnose DGM deficiencies.}

%% file: figures/background_tables/background_table.tex
\begin{table*}[t]
    \centering
    \begin{tabular}{l| l l c c c}
    \hline
        Name & Objective & Divergence & $\pi_i > 0$ & $\pi_i = 0$ & $\pi_i < 0$ \\
        \hline
        Empirical Likelihood & $\prod_i \pi_i$ & $\klkl{\hat{P}_n}{P_{\boldsymbol{\pi}}}$ & Y & N & N \\
        Exponential Tilting & $-\sum_i \pi_i \log \pi_i$ & $\klkl{P_{\boldsymbol{\pi}}}{\hat{P}_n}$ & Y & Y & N \\
        Euclidean Likelihood & $\frac{1}{2}\sum_i (\pi_i - n^{-1})^2$ & $\frac{1}{2}\sum_i (\pi_i - n^{-1})^2$ & Y & Y & Y \\
    \end{tabular}
    \resizebox{\textwidth}{!}{\begin{tabular}{l| c c c c}
        \hline
        Name & Population Statistic $S$ &Condition at $S=0$ & Equivalent $\vm(\X; \vc)$ & $\vc$\\
        \hline
        Mean & $(\mu_x - \vc)^{\top}(\mu_x - \vc)$ & $\mu_x = \vc$ & $\X - \vc$ & $\vc$\\
        Score Function & $\|\E[\nabla_{\theta} \log p_{\theta}(x)]\|^2$ & Maximum Likelihood & $\nabla_{\theta} \log p_{\theta}(\suman{\X})$  & $\mathbf{0}$\\
        \fid & $\| \mu_x - \mu_y \|^2 + \mathrm{tr}(\Sigma_x + \Sigma_y - 2(\Sigma_x \Sigma_y)^{1/2})$ & $\mu_x = \mu_y, \Sigma_x = \Sigma_y$ & $[\phi(\X), \phi(\X)\phi(\X)^{\top}] - \vc$ & $\E_y[[\phi(\Y), \phi(\Y)\phi(\Y)^{\top}]]$\\
        Mean Embedding & $\sum_i (\E_x[k(x, t_i)] - \E_y[k(y, t_i)])^2$ & \suman{$P=Q \implies S=0$ a.s., $P \neq Q \implies S\neq0$ a.s.} & $k(x, t_i)-k(y, t_i) \quad i=1,\dots,W$ & $\mathbf{0}$\\
    \end{tabular}}
    \caption{Top: Generalized Empirical Likelihood objectives and valid values of $\pi_i$. Bottom: Common statistics as moment conditions.}
    \vspace{-3mm}
    \label{tab:EL_background}
\end{table*}

%% file: 03_gen_models_tests.tex
\section{GEL for Evaluating Generative Models}
\label{sec:dgm_tests}
\looseness=-1 \textbf{Setup} We assume that we have samples $x_i$ from the data distribution $p$, and access to a DGM with distribution $q$, from which we can generate samples $y_j$.
In the GEL approach to generative model evaluation, we make three choices: the constant $\vc$, which is a function of the generative model distribution \suman{$q$} and embeds relevant information about the DGM; the associated moment function $\vm(\X; \vc)$, whose expectation equals $\mathbf{0}$ should \suman{relevant information about} $p$ and $q$ match; and the divergence $\kl{\hat{P}_n}{P_{\pi}}$, \suman{which forms the objective}. %
Before \suman{describing} the proposed \suman{metrics}, we provide a motivating example. %
\label{sec:gen_models}
\subsection{A Motivating Example}
\label{ssec:motivating_example}
\input{figures/mode_drop_table_v5}
\looseness=-1 To illustrate the utility of GEL method for DGM evaluation, we create a simple test that identifies \emph{which} modes the generative model has dropped. \suman{Ours is a GEL test of} the mean: $\vc = \E_{y\sim q}[\phi(y)]$ is the mean of a feature function $\phi$, $\vm(\X; \vc) = \phi(\X) - \vc$ is the moment function, and the reverse KL $\klkl{P_{\boldsymbol{\pi}}}{\hat{P}_n}$ is the divergence. The resulting problem is an exponentially tilting GEL.

The test requires a fourth choice, an appropriate feature space $\phi$. The key idea for choosing $\phi$ is that if a test set contains samples $x$ from a missing mode, $\phi$ must be designed such that the weight $\pi_i$ of feature $\phi(x_i)$ goes to zero. The following lemma offers some insight into one approach.
\begin{restatable}{lemma}{el_orthogonality}
\label{thm:el_orthogonality}
Assume that the true data distribution $p$ is a mixture of the model distribution $q$ and another distribution $\od$. We consider an estimation $\hat{\E}_q [\phi(y)]$ of the model mean obtained using samples $y_i, i \in \mathcal{I} \subset \mathbb{N}$ from $q$; and a test set composed of samples from $p$. The test set can be split into samples $\{\chuck_1, \dots, \chuck_m \}\sim \od$ and samples $\{\keep_{m+1}, \dots, \keep_{n}\}\sim q$. Then, the mean equality condition is 
\begin{align}
\hat{\E}_{q} [\phi(y)] =  \sum_{i=1}^m \pi_i \phi(\chuck_i) + \sum_{i=m+1}^n \pi_i \phi(\keep_i)%
\label{eq:el_ort_mean_condition}
\end{align}
If the convex hull $\mathrm{Conv} \{ \phi(\chuck_i), i=1,\dots,m \}$ does not intersect with $ \mathrm{Span} \{ \phi(y_i), i \in \mathcal{I} \cup \{m+1,\dots,n \} \}$ then $\pi_i=0, i=1, \dots, m$.
\end{restatable}

The proof can be found in \cref{sec:app_proofs}. %
Note that deep features, such as those from the Pool3 layer of the Inception v3 network \citep{szegedy2016rethinking}, the FC2 layer of the VGG16 network \citep{simonyan2014very}, or BYOL features \citep{grill2020bootstrap}, approximately satisfy these properties. As features lie in the non-negative orthant (as it is an output of ReLU), $\mathbf{0}$ does not lie within the convex hull of any set of points.
Since the features of different classes are approximately linearly separable, it is unlikely that linear combination of features of a dropped mode would be part of the solution that satisfies the mean constraint. %

In addition to the features, we chose the reverse KL divergence as our objective, since by construction we require $\pi_i$ to be $0$ for certain test set samples. Note that we need not worry about the normalization of $\phi(x)$, since we obtain the same $\pi_i$ for any full-rank linear transformation of $\phi$. %

\emph{Experiment:} To illustrate the performance of the method when modes are missing, we conduct a controlled study. We construct synthetic generative models from the CIFAR-10 training set \citep{krizhevsky2009learning}. Each of these ``generative models'' contains 5,000 total examples per class, with a varying number of missing classes (ranging from 0 to 8). We use the CIFAR-10 test set as the data distribution. %

\looseness=-1 As shown in \cref{fig:mode_drop}, \suman{GEL} weights %
are highly sensitive to which modes were dropped, and can be used as \suman{a} diagnostic tool for evaluating models.\footnote{Empirical likelihoods are also highly sensitive, but they also end up at the boundary of the convex hull after a few modes are dropped.} We note that our approach is sensitive to \emph{which} modes were dropped even though the features used were not trained on CIFAR-10 data. We also compare the method to the improved recall metric \citep{kynkaanniemi2019improved}, and per-class coverage probabilities \citep{ferjad2020reliable} (the latter with four nearest neighbors, as parameter gave us the recommended $.95$ coverage for 0 dropped modes). The former substantially overestimates the probability of the missing mode.  The latter, while performing better than recall, overestimates probabilities for multiple missing modes. %
GEL \suman{outperforms both and moreover, }does not require tuning of the highly sensitive nearest neighbor parameter $k$. Finally, as shown on the right-hand side of \cref{fig:mode_drop}, GEL is not hugely reliant on the Pool3 space: it performs similarly on Bootstrap Your Own Latent (BYOL) features. 
\input{03b_proposed_tests}

\input{03c_two_sample_model_comp}

\input{figures/validation_experiments/unbalanced_mode_with_kgel_drop}
\looseness=-1 \suman{\textbf{Calculation} We solve GEL objectives using Newton's method on the dual problem. The computational complexity is $\mathcal{O}(nd^3)$, where $n$ and $d$ are the number of samples and dimensionality, respectively. For detailed derivations of the dual and for code, we refer the reader to \cref{sec:app_calculation}. An implementation is available at \url{https://github.com/deepmind/understanding_deep_generative_models_with_generalized_empirical_likelihood/}}

%% file: figures/mode_drop_table_v5.tex
\newcommand{\tabsetmedium}{\setlength{\tabcolsep}{12pt}}
\newcommand{\tablesize}{\scriptsize}
\newcommand{\droppedmodehellingertablekernel}{
    \bgroup
    \setlength{\tabcolsep}{2pt}
    \renewcommand{\arraystretch}{0.85}
    \begin{tabular}{c l c c c c c}
    & \scriptsize{No. Missing Modes} & \tablesize 0 & \tablesize 2 & \tablesize 4 & \tablesize 6 & \tablesize 8 \\
    \midrule
    \multirow{3}{*}{\rotatebox[origin=c]{90}{\tablesize{Pool3}}} & \tablesize{64 Witness Points} & \tablesize 0.0058 & \tablesize 0.1891 & \tablesize 0.3057 & \tablesize 0.3983 & \tablesize 0.3630 \\
    & \tablesize{256 Witness points} & \tablesize 0.0064 & \tablesize 0.1682 & \tablesize 0.2739 & \tablesize 0.3528 & \tablesize 0.3197 \\
    & \tablesize{1024 Witness points} & \tablesize 0.0074 & \tablesize 0.1423 & \tablesize 0.2390 & \tablesize 0.3060 & \tablesize 0.2750 \\
    \bottomrule
    \end{tabular}
    \egroup
}
\newcommand{\droppedmodehellingertable}{
    \bgroup
    \setlength{\tabcolsep}{2pt}
    \renewcommand{\arraystretch}{0.85}
    \begin{tabular}{c l c c c c c}
    & \scriptsize{No. Missing Modes} & \tablesize 0 & \tablesize 2 & \tablesize 4 & \tablesize 6 & \tablesize 8 \\
    \midrule
    \multirow{4}{*}{\rotatebox[origin=c]{90}{\tablesize{Pool3}}} & \tablesize Chance & \tablesize \textbf{0.0000} & \tablesize 0.3249 & \tablesize 0.4748 & \tablesize 0.6063 & \tablesize 0.7435 \\
    & \tablesize{Improved Recall} & \tablesize 0.0174 & \tablesize 0.3068 & \tablesize 0.4368 & \tablesize 0.5712 & \tablesize 0.6401 \\
    & \tablesize{Coverage} & \tablesize 0.0021 & \tablesize 0.1582 & \tablesize 0.3154 & \tablesize 0.4142 & \tablesize 0.4811 \\
    & \textbf{\tablesize{GEL (Ours)}} & \tablesize 0.0092 & \tablesize \textbf{0.1427} & \tablesize \textbf{0.2360} & \tablesize \textbf{0.3059} & \tablesize \textbf{0.3670} \\
    \bottomrule
    \end{tabular}
    \egroup
}
\newcommand{\droppedmodehellingertablebyol}{
    \bgroup
    \setlength{\tabcolsep}{2pt}
    \renewcommand{\arraystretch}{0.85}
    \begin{tabular}{c l c c c c c}
     & \scriptsize{No. Missing Modes} & \tablesize 0 & \tablesize 2 & \tablesize 4 & \tablesize 6 & \tablesize 8 \\
    \midrule
    \multirow{4}{*}{\rotatebox[origin=c]{90}{\tablesize{BYOL}}} & \tablesize Chance & \tablesize \textbf{0.0000} & \tablesize 0.3249 & \tablesize 0.4748 & \tablesize 0.6063 & \tablesize 0.7435 \\
    & \tablesize{Improved Recall} & \tablesize 0.0154 & \tablesize 0.2070 & \tablesize 0.3625 & \tablesize 0.4988 & \tablesize 0.5094 \\
    & \tablesize{Coverage} & \tablesize 0.0023 & \tablesize 0.2196 & \tablesize 0.3712 & \tablesize 0.4749 & \tablesize 0.5588 \\
    & \textbf{\tablesize{GEL (Ours)}} & \tablesize 0.0055 & \tablesize \textbf{0.1465} & \tablesize \textbf{0.2460} & \tablesize \textbf{0.3122} & \tablesize \textbf{0.3791} \\
    \bottomrule
    \end{tabular}
    \egroup
}
\newcommand{\droppedmodehellingertabletranspose}{
    \bgroup
    \setlength{\tabcolsep}{2pt}
    \renewcommand{\arraystretch}{0.95}
    \begin{tabular}{c 1| c c c}
    \toprule
    & & \rotatetext{\tablesize Impr. Recall} & \rotatetext{\tablesize Coverage} & \rotatetext{\tablesize GEL (Ours)} \\
    \midrule
    \multirow{5}{*}{\rotatetext{\scriptsize{\# of Missing Modes}}} 
    &  \tablesize 0 & \tablesize 0.0135 & \tablesize \textbf{0.0038} & \tablesize 0.0092 \\
    & \tablesize 2 & \tablesize 0.3162 & \tablesize 0.1504 & \tablesize \textbf{0.1427} \\
    & \tablesize 4 & \tablesize 0.4534 & \tablesize 0.2852 & \tablesize \textbf{0.2360} \\
    & \tablesize 6 & \tablesize 0.5861 & \tablesize 0.3949 & \tablesize \textbf{0.3059} \\
    & \tablesize 8 & \tablesize 0.6619 & \tablesize 0.4650 & \tablesize \textbf{0.3670} \\
    \bottomrule
    \end{tabular}
    \egroup
    \tabsetmedium
}
\newcommand{\insertmodedropeventshort}[8]{
    \multicolumn{2}{c}{\rotatebox[origin=c]{90}{\footnotesize{#1}}} &
    \begin{minipage}{0.18\linewidth}
      \includegraphics[width=\linewidth, trim=0 0 0 0, clip]{figures/10k_egs_mode_drops/drop_mode_#3_#4_#2_fig_v3.pdf}
    \end{minipage}
    &
    \begin{minipage}{0.18\linewidth}
      \includegraphics[width=\linewidth, trim=0 0 0 0, clip]{figures/10k_egs_mode_drops/drop_mode_#3_#5_#2_fig_v3.pdf}
    \end{minipage}
    &
    \begin{minipage}{0.18\linewidth}
      \includegraphics[width=\linewidth, trim=0 0 0 0, clip]{figures/10k_egs_mode_drops/drop_mode_#3_#6_#2_fig_v3.pdf}
    \end{minipage}
    &
    \begin{minipage}{0.18\linewidth}
      \includegraphics[width=\linewidth, trim=0 0 0 0, clip]{figures/10k_egs_mode_drops/drop_mode_#3_#7_#2_fig_v3.pdf}
    \end{minipage}
    &
    \begin{minipage}{0.18\linewidth}
      \includegraphics[width=\linewidth, trim=0 0 0 0, clip]{figures/10k_egs_mode_drops/drop_mode_#3_#8_#2_fig_v3.pdf} 
    \end{minipage}
    \\
}
\newcommand{\insertmodedropeventlong}[9]{
    \rotatebox[origin=c]{90}{\footnotesize{#1}} &
    \rotatebox[origin=c]{90}{\footnotesize{#2}} &
    \begin{minipage}{0.18\linewidth}
      \includegraphics[width=\linewidth, trim=0 0 0 0, clip]{figures/10k_egs_mode_drops/drop_mode_#4_#5_#3_fig_v3.pdf}
    \end{minipage}
    &
    \begin{minipage}{0.18\linewidth}
      \includegraphics[width=\linewidth, trim=0 0 0 0, clip]{figures/10k_egs_mode_drops/drop_mode_#4_#6_#3_fig_v3.pdf}
    \end{minipage}
    &
    \begin{minipage}{0.18\linewidth}
      \includegraphics[width=\linewidth, trim=0 0 0 0, clip]{figures/10k_egs_mode_drops/drop_mode_#4_#7_#3_fig_v3.pdf}
    \end{minipage}
    &
    \begin{minipage}{0.18\linewidth}
      \includegraphics[width=\linewidth, trim=0 0 0 0, clip]{figures/10k_egs_mode_drops/drop_mode_#4_#8_#3_fig_v3.pdf} 
    \end{minipage}
    &
    \begin{minipage}{0.18\linewidth}
      \includegraphics[width=\linewidth, trim=0 0 0 0, clip]{figures/10k_egs_mode_drops/drop_mode_#4_#9_#3_fig_v3.pdf} 
    \end{minipage}
    \\
}
\newcommand{\modedropfiftyk}{
    \includegraphics[width=0.98\linewidth, trim=0 0 0 0, clip]{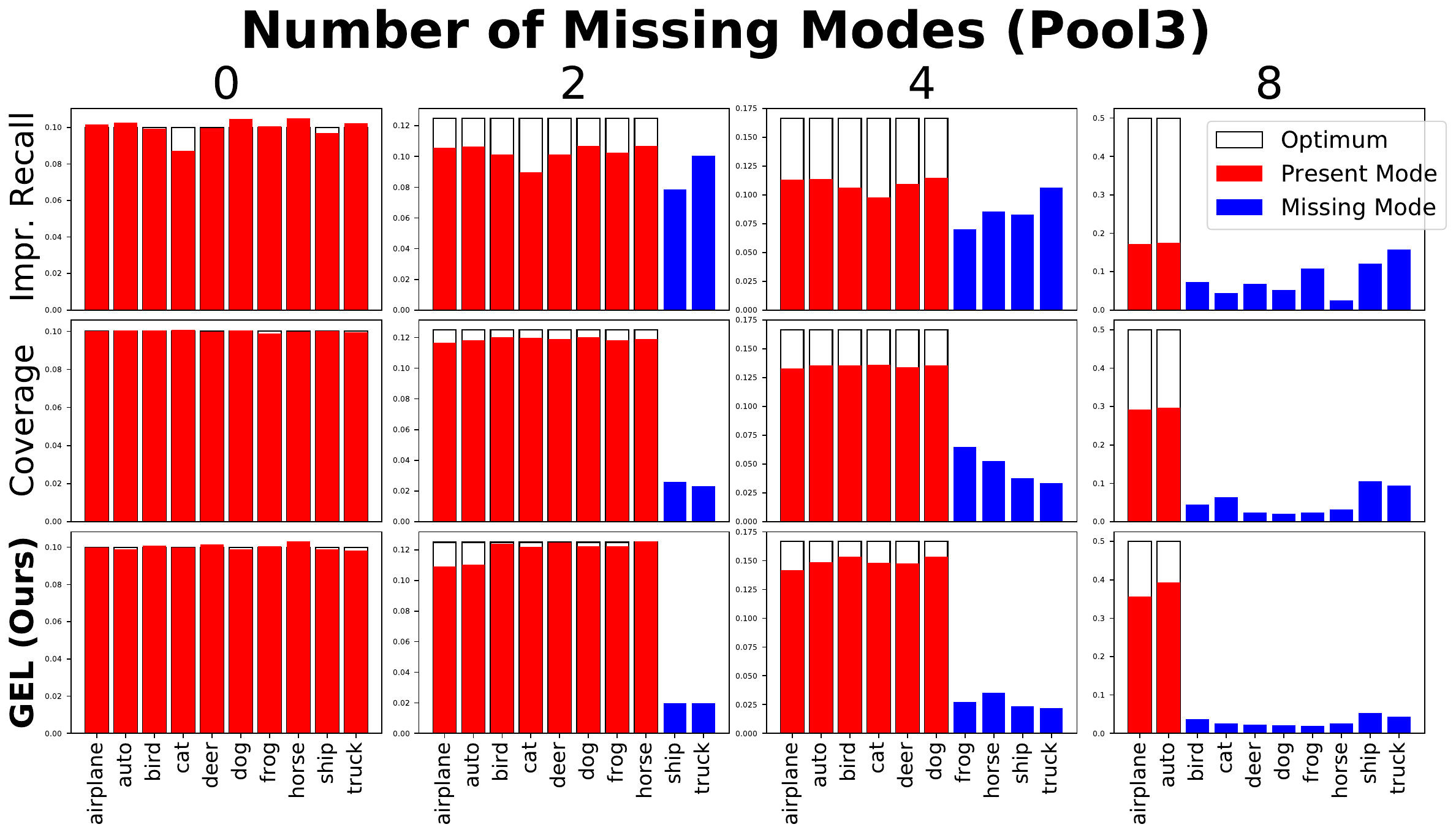}
}
\newcommand{\modedropfiftykbyol}{
    \includegraphics[width=0.98\linewidth, trim=0 0 0 0, clip]{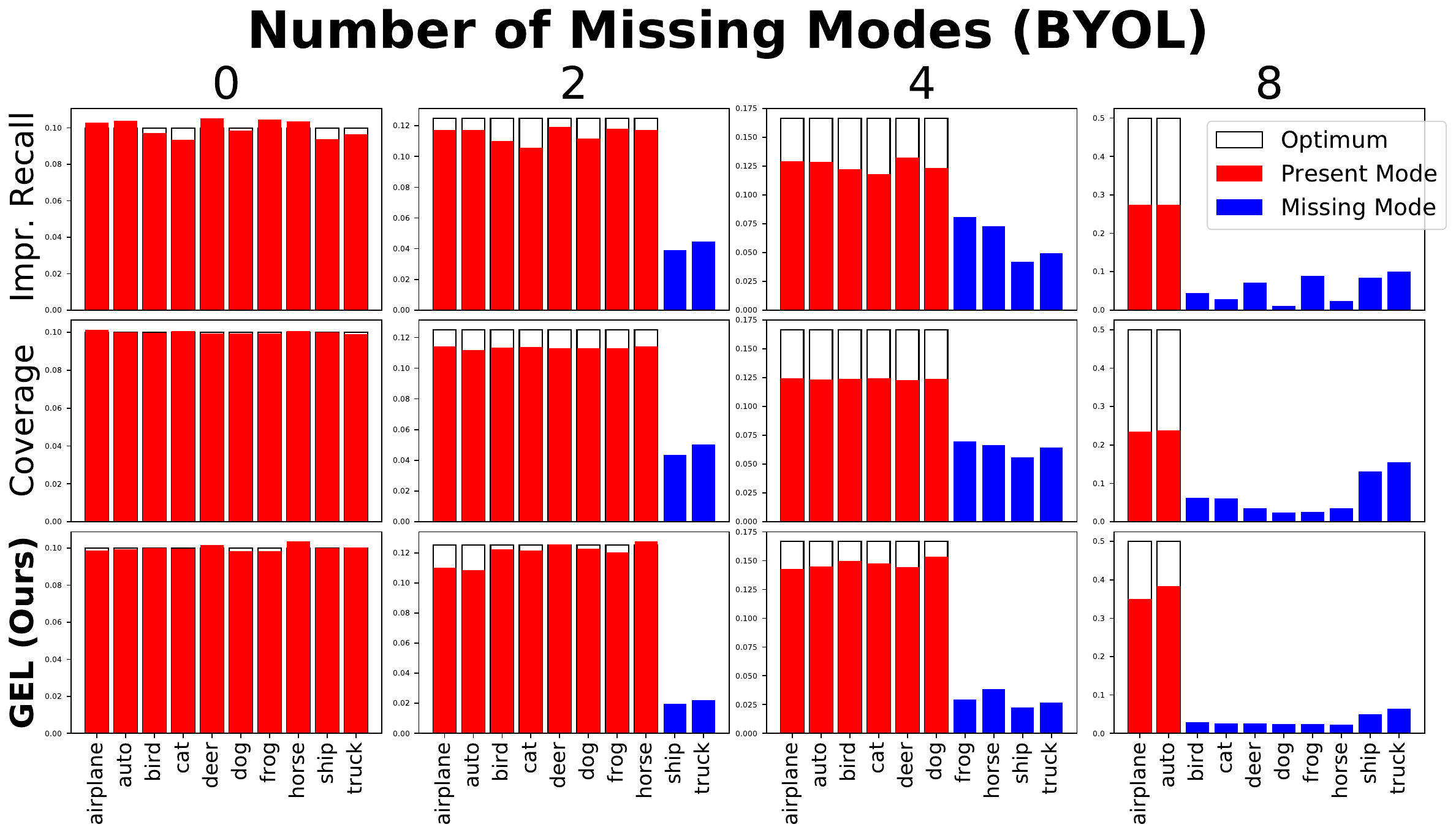}
}
\newcommand{\datasetkernelmodedrop}{
    \setlength{\tabcolsep}{1pt}
    \begin{tabular}{c c c c c c c}
        \toprule
        && \multicolumn{5}{c}{\textbf{Number of Missing Modes KGEL} } \\
        & & 0 & 2 & 4 & 6 & 8 \\
        \midrule
        \insertmodedropeventlong{64}{ witness}{pool3_ker_gel_poly3_64_wit}{real}{0}{2}{4}{6}{8}
        \insertmodedropeventlong{256}{ witness}{pool3_ker_gel_poly3_256_wit}{real}{0}{2}{4}{6}{8}
        \insertmodedropeventlong{1024}{ witness}{pool3_ker_gel_poly3_1024_wit}{real}{0}{2}{4}{6}{8}
        \bottomrule
    \end{tabular}
    \tabsetmedium
}
\newcommand{\datasetmodedrop}{
    \setlength{\tabcolsep}{1pt}
    \begin{tabular}{c c c c c c c}
        \toprule
        && \multicolumn{5}{c}{\textbf{Number of Missing Modes (Pool3)}} \\
        & & 0 & 2 & 4 & 6 & 8 \\
        \midrule
        \insertmodedropeventlong{Impr.}{Recall}{pool3_knn_5_norm_recall}{real}{0}{2}{4}{6}{8}
        \insertmodedropeventlong{Cove-}{rage}{pool3_knn_4_norm_coverage}{real}{0}{2}{4}{6}{8}
        \insertmodedropeventlong{\textbf{GEL}}{ (Ours)}{pool3_gel}{real}{0}{2}{4}{6}{8}
        \bottomrule
    \end{tabular}
    \tabsetmedium
}
\newcommand{\datasetbyolmodedrop}{
    \setlength{\tabcolsep}{1pt}
    \begin{tabular}{c c c c c c c}
        \toprule
        && \multicolumn{5}{c}{\textbf{Number of Missing Modes (BYOL)}} \\
        & & 0 & 2 & 4 & 6 & 8 \\
        \midrule
        \insertmodedropeventlong{Impr.}{Recall}{byol_knn_5_norm_recall}{real}{0}{2}{4}{6}{8}
        \insertmodedropeventlong{Cove-}{rage}{byol_knn_4_norm_coverage}{real}{0}{2}{4}{6}{8}
        \insertmodedropeventlong{\textbf{GEL}}{ (Ours)}{byol_gel}{real}{0}{2}{4}{6}{8}
        \bottomrule
    \end{tabular}
    \tabsetmedium
}
\newcommand{\rotatetext}[1]{
    \rotatebox[origin=c]{90}{#1}
}
\begin{figure*}[t]
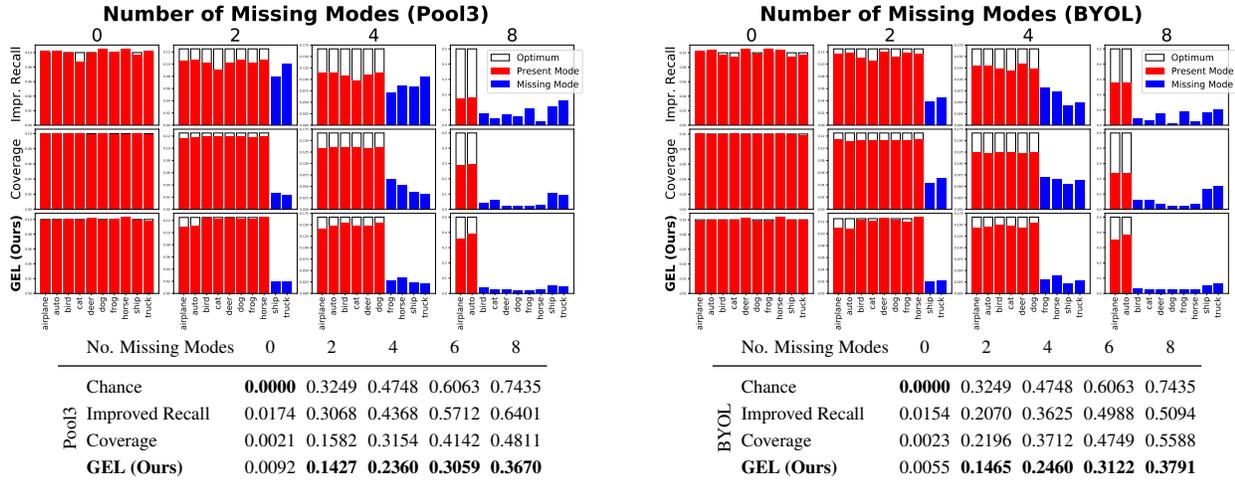

\centering

\renewcommand{\arraystretch}{1.25}
\tabsetmedium
\begin{tabular}{cc}
    \begin{minipage}{0.45\linewidth}
        \modedropfiftyk
    \end{minipage} &
    \begin{minipage}{0.45\linewidth}
        \modedropfiftykbyol
     \end{minipage}\\
    \begin{minipage}{0.45\linewidth}
        \centering
        \droppedmodehellingertable%
     \end{minipage} &
     \begin{minipage}{0.45\linewidth}
     \centering
     \droppedmodehellingertablebyol
     \end{minipage}
\end{tabular}
\caption{\looseness=-1 Mode dropping detection on CIFAR-10 data using Pool3 features (Left panel) or BYOL features (Right panel). We vary the number of missing modes in the training set and compare three different metrics computed on the test set: aggregate improved recall ``weights'' (Top row),  aggregate Coverage ``weights''(Second) and GEL weights (Third). Weights in the figures rescaled such that the optimal weight for a  mode present in the training set is 1.
Tables at the bottom show the Hellinger distance between ground truth probabilities and weights obtained using the three metrics.}
\label{fig:mode_drop}
\end{figure*}
\setlength{\tabcolsep}{6pt}
\renewcommand{\arraystretch}{1.0}

%% file: 03b_proposed_tests.tex
\subsection{Proposed Tests}
\label{ssec:proposed_test}
We extend the test in \cref{ssec:motivating_example} in various ways to improve upon the diagnostic accuracy of the metric, and to make it more broadly applicable for generative model evaluation. We focus on three specific areas. First, we use ideas from MMD and ME to create GEL test of distributions. Second, we further use ideas from kernelization to include label information in the constraint, creating a test for conditional generative models. Finally, to make the metric more robust to model misspecification, improve interpretability of the metric, and ensure model comparison, we introduce a two-sample test and a modification of the one-sample test.

\looseness=-1 \textbf{A Distribution Test by combining GEL and ME}
If the test in \cref{ssec:motivating_example} is the test of the mean, then by analogy, \suman{one only needs to replace the mean condition to create GEL tests of distributions. We use moment conditions based on Mean Embeddings (ME) referenced in \cref{sec:gel_intro} for the appropriate moment restrictions, which importantly are linear in $\boldsymbol{\pi}$}.

\looseness=-1 To \suman{render} the test useful for DGM evaluation, we make three design choices. First, similar to the mean test in \cref{ssec:motivating_example}, we use a feature space. Second, to improve interpretability, we use multiple witness points (typically 50--1,024), which are sampled from a validation set of the data distribution.\footnote{Technically, this violates \suman{the condition that the witness distribution be absolutely continuous with respect to the Lebesgue measure}. We could easily fix the violation by adding a small amount of Gaussian noise to the witness points. This, however, makes no practical difference \suman{to the resulting GEL solution.}} Finally, we use the exponential kernel $k(x, y) = \exp(x^\intercal y /d )$ (where $d$ is the dimension of the vectors), which is analytic and characteristic \suman{on compact sets of $\mathbb{R}^d$  \cite{muandet2016kernel}}.\suman{\footnote{Also, the kernel is trivially integrable on compact sets of $\mathbb{R}^d$.}} This gives us the \textbf{Kernel GEL (\KGEL)} test
\begin{equation}
\label{eq:kgel}
\begin{gathered}
\min_{\{\boldsymbol{\pi} | \sum_i \pi_i=1, \pi_i \geq 0\}} \klkl{P_{\boldsymbol{\pi}}}{\hat{P}_n} \quad \mbox{subject to} \\
\sum_{i=1}^n \pi_i 
{\scriptscriptstyle \begin{bmatrix}
k(\phi(x_i), \phi(t_1))\\
\vdots \\
k(\phi(x_i), \phi(t_W))
\end{bmatrix}}
=
\E_{y\sim q} \begin{bmatrix}
k(\phi(y), \phi(t_1))\\
\vdots \\
k(\phi(y), \phi(t_W))
\end{bmatrix}.%
\end{gathered}
\end{equation}

\textbf{Including Label Information via Kernelization}
A second benefit of kernelization is that it allows us to easily include label information in the metric, enabling us to  create tests for conditional generative models. The extension is straightforward: if we consider $x = (x^{(d)}, x^{(l)})$ to include \suman{both} the image $x^{(d)}$ and its associated label $x^{(l)}$, respectively, then we can \suman{construct} kernels $k(x, t)$ that include both image and label information. We use the product kernel $k(x, t) = k_d(x^{(d)}, t^{(d)}) k_l(x^{(l)}, t^{(l)})$, where $k_d$ and $k_l$ are image and label kernels, respectively. If $k_d$ and $k_l$ are universal, then the product kernel is also universal \suman{and therefore characteristic} \cite{szabo2017characteristic}. A test on the joint distribution is simply \cref{eq:kgel} with the appropriate product kernel.\footnote{As we use exponential kernel, which \suman{is} characteristic but not universal, the product kernel may not be characteristic. We find this distinction to be more of a theoretical than practical concern.}

\looseness=-1 The choice of label kernel depends on the type of label information. If the dataset contains a small number of labels, the delta kernel, $k_l(x^{(l)}, t^{(l)}) = \mathbb{I}_{[x^{(l)} = t^{(l)}]}$ is an appropriate choice (as it is universal \cite{muandet2016kernel}). $k_l$ assigns $0$ similarity to two points that do not have the same label. We use this test in \cref{ssec:validation} to \suman{identify samples where DGMs ignore label information.}

When a label hierarchy is available, such as for ImageNet, %
\suman{we can construct a kernel that measures similarity between labels in a more fine-grained way}. %
\suman{First,} we associate a label with the path from the root node to the label leaf node, encoding it as a string. We then compute similarity between labels using any string kernel, with one simple option being the special case of the Smith-Waterman score using no gap penalty and an identity substitution matrix \cite{vert2004kernel}. \suman{We provide further details in Appendix D.}

%% file: 03c_two_sample_model_comp.tex
\looseness=-1 \textbf{Accounting for Model Misspecification with Two-sample Tests}
 DGMs may not only fail to represent part of the data distribution, but also may generate samples %
 \suman{outside of} the support of the data distribution. One can perform the previously described tests by letting $\vc$ be a function of data, and \suman{$\X$ DGM samples.} There exists, however, a \suman{larger} issue: if the DGM produces many \suman{out of distribution examples} (or a single example sufficiently out of distribution), then there may not exist \suman{a $\boldsymbol{\pi}$} that \suman{satisfies} the moment restriction.\footnote{\suman{For mean tests, \suman{this is geometrically equivalent to } %
 the model mean \suman{not lying in} the convex hull of test points. Again, due to space constraints, we defer discussion of the convex hull condition to \cref{ssec:convex_hull_condition}.}}

\looseness=-1 We address these issues by \suman{using} a two-sample extension to GEL \citep{qin1994semi}.
In this version, we assign \suman{a first set of}  probabilities $\pi_i$ to test points $x_i$, and a second set of probabilities $\psi_j$ to model points $y_j$. 
The two-sample variant of \suman{the} mean test, which we \suman{denote} \textbf{Two-sample Generalized Empirical Likelihood (GEL2)} is
 \begin{equation}
 \begin{gathered}
 \min_{\boldsymbol{\pi}, \boldsymbol{\psi} \geq \boldsymbol{0}}  \kl{\hat{P}_n}{P_{\boldsymbol{\pi}}} + \kl{\hat{P}_m}{P_{\boldsymbol{\psi}}} %
 ~~\mbox{subject to}~~ \\
     \sum_{i=1}^n \pi_i \phi(x_i) = \sum_{j=1}^m \psi_j \phi(y_j), \; \sum_{i=1}^n \pi_i = \sum_{j=1}^m \psi_j = 1.
\end{gathered}
\end{equation}
\suman{The two-sample test finds, among distributions that satisfy the moment conditions, the two that are closest to their respective empirical distributions.} We call the ME version the \textbf{Kernel Two-sample Generalized Empirical Likelihood (\ktsgel)}.
This variant improves upon the one-sample test in two ways. %
First, it opens up more diagnostic tools: we can find DGM samples that are not in the support of the data distribution (\cref{ssec:real_world}). Second, it provides finite scores for a wider range of generative models: the intersection of the two convex hulls only needs to be non-empty.

\looseness=-1 The intersection of the convex hulls may be empty, in which case the empirical likelihood is $-\infty$. In practice this only occurs with particularly poorly performing generative models, in which case the per-sample probabilities are unlikely to be helpful. %

%% file: figures/validation_experiments/unbalanced_mode_with_kgel_drop.tex
\renewcommand{\tablesize}{\scriptsize}
\newcommand{\unbalancedmodetabular}{
    \bgroup
    \renewcommand{\arraystretch}{0.85}
    \resizebox{\textwidth}{!}{\begin{tabular}{c|c c c c}
        \toprule
        Prob. & \multirow{2}{*}{Chance} & Improved & \multirow{2}{*}{Coverage} & \textbf{\KGEL} \\
        Mode &   & Recall   &                           &   (Ours)   \\
        \midrule
          0.0 & 0.5412 & 0.4481 & 0.4262 & \textbf{0.2789} \\ %
          0.1 & 0.3249 & 0.3016 & 0.3036 & \textbf{0.1178} \\ %
         0.3 & 0.1452 & 0.1329 & 0.1421 & \textbf{0.0444} \\ %
         0.5 & \textbf{0.0000} & 0.0073 & \textbf{0.0000} & 0.0021 \\ %
         0.7 & 0.1452 & 0.1484 & 0.1419 & \textbf{0.0473} \\ %
         0.9 & 0.3249 & 0.3238 & 0.3000 & \textbf{0.1157} \\ %
        1.0 & 0.5412 & 0.4994 & 0.3939 & \textbf{0.2891} \\ %
        \bottomrule
    \end{tabular}}
    \egroup
}
\newcommand{\unbalancedmodetabularfortyk}{
    \bgroup
    \renewcommand{\arraystretch}{0.85}
    \resizebox{\textwidth}{!}{\begin{tabular}{c|c c c c}
        Mode 1 & \multirow{2}{*}{Chance} & Improved & \multirow{2}{*}{Coverage} & \textbf{\KGEL} \\
        Probability &   & Recall   &                           &   (Ours)   \\
        \midrule
          0.1 & 0.3249 & 0.3021 & 0.2808 & \textbf{0.1206} \\ %
         0.3 & 0.1452 & 0.1313 & 0.1354 & \textbf{0.0486} \\ %
         0.5 & \textbf{0.0000} & 0.0080 & 0.0007 & 0.0030 \\ %
         0.7 & 0.1452 & 0.1506 & 0.1323 & \textbf{0.0434} \\ %
         0.9 & 0.3249 & 0.3238 & 0.2706 & \textbf{0.1207} \\ %
    \end{tabular}}
    \egroup
}
\newcommand{\unbalancedmodecomparison}{
    \includegraphics[width=0.98\linewidth, trim=0 0 0 0, clip]{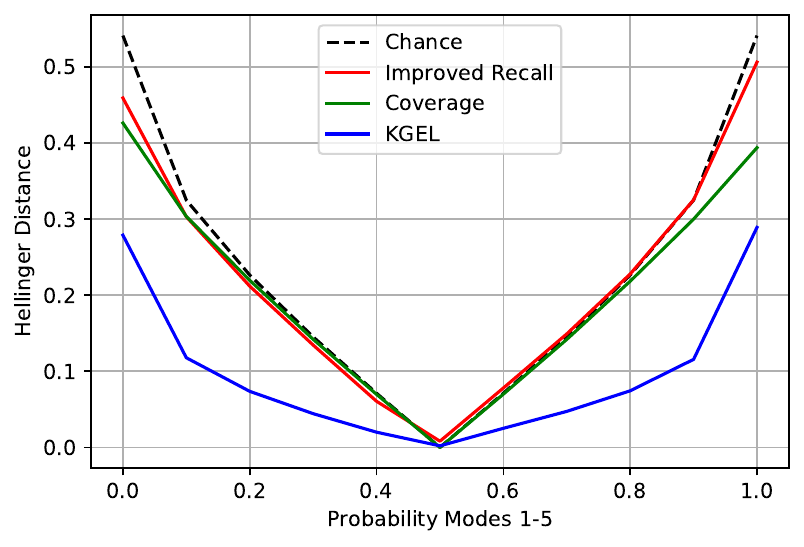}
}
\newcommand{\unbalancedmodecomparisonfortyk}{
    \includegraphics[width=0.96\linewidth, trim=0 0 0 0, clip]{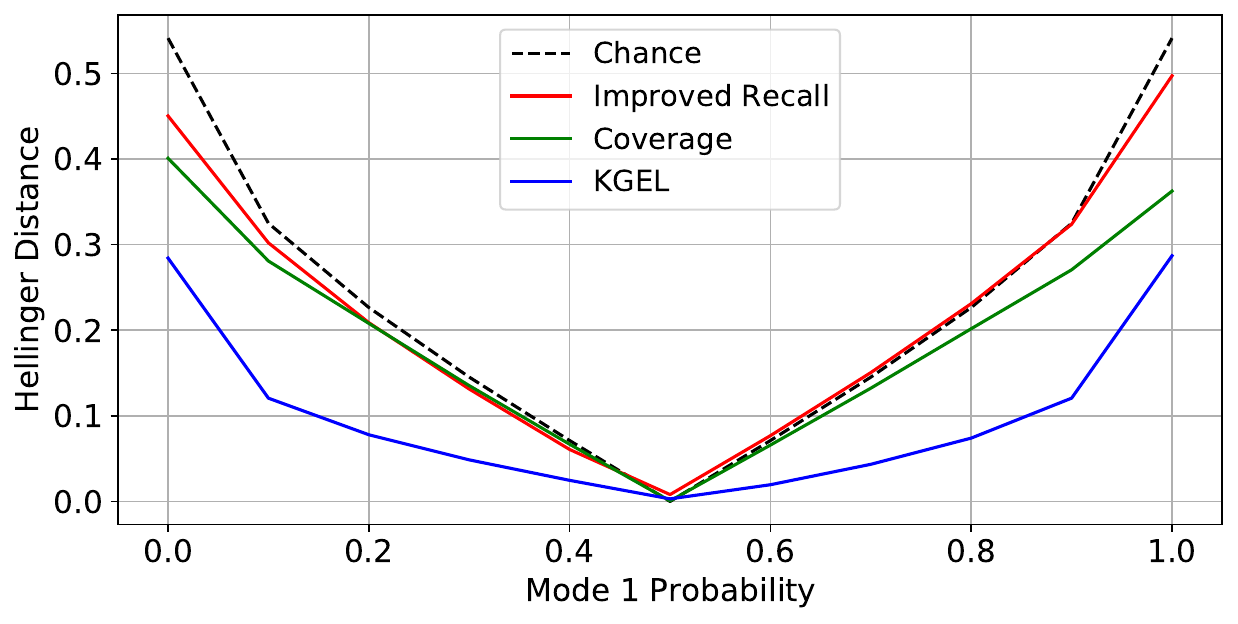}
}
\newcommand{\droppedmodecomparisonfortykold}{
    \includegraphics[width=\linewidth, trim=20 0 80 0, clip]{figures/unbalanced_modes/mode_drop_40k_0_4_8.pdf}
}
\newcommand{\droppedmodecomparisonfortyk}{
    \includegraphics[width=\linewidth, trim=0 0 0 0, clip]{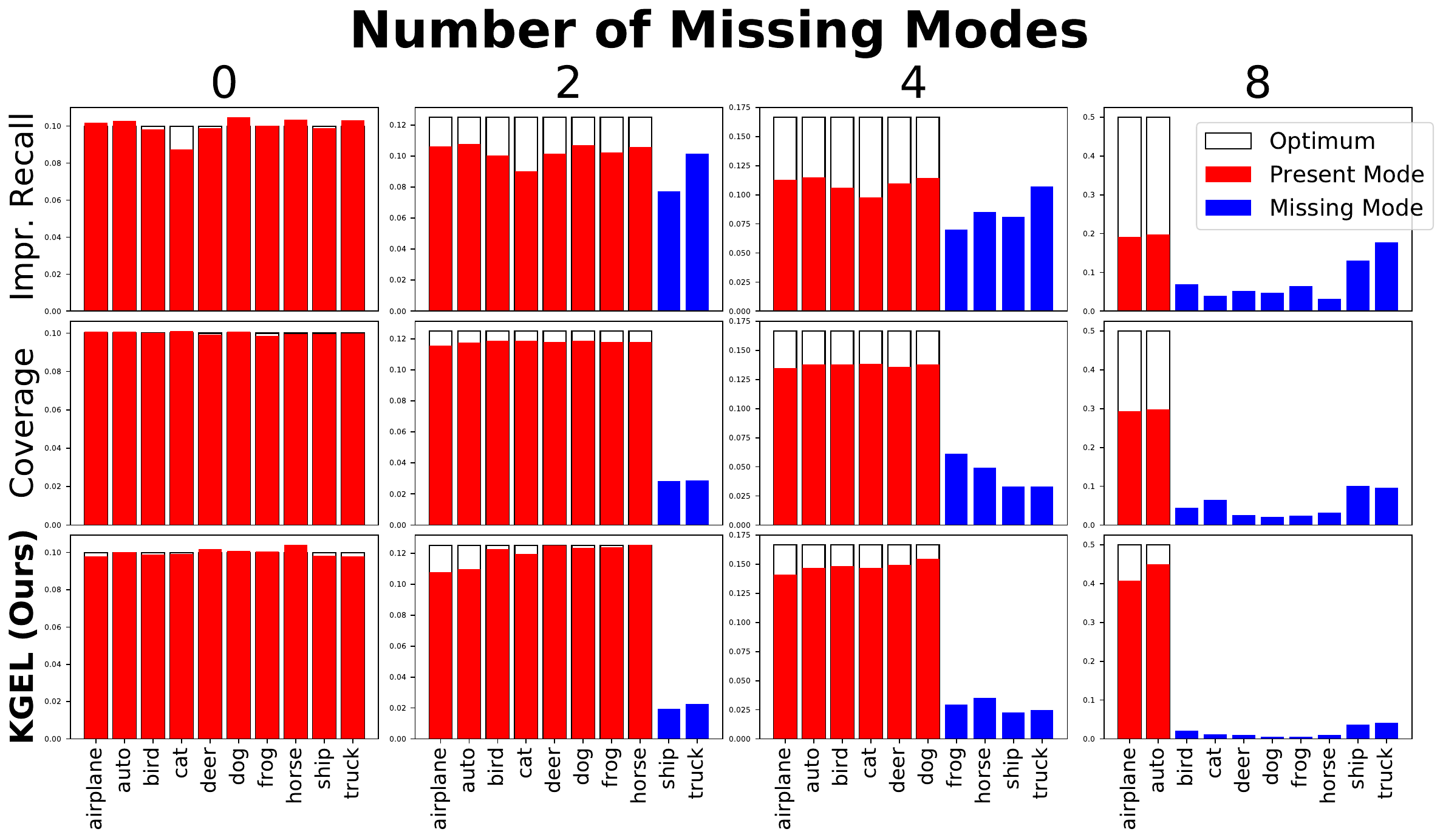}
}
\newcommand{\droppedmodehellingertablekernelfortyk}{
    \bgroup
    \setlength{\tabcolsep}{2pt}
    \renewcommand{\arraystretch}{0.85}
    \begin{tabular}{c l c c c c c}
    & \scriptsize{No. Missing Modes} & \tablesize 0 & \tablesize 2 & \tablesize 4 & \tablesize 6 & \tablesize 8 \\
    \midrule
    \multirow{3}{*}{\rotatebox[origin=c]{90}{\tablesize{Pool3}}} & \tablesize{Nvidia Recall} & \tablesize  & \tablesize  & \tablesize  & \tablesize  & \tablesize  \\
    & \tablesize{Coverage} & \tablesize \textbf{} & \tablesize  & \tablesize  & \tablesize  & \tablesize  \\
    & \textbf{\tablesize{\KGEL (Ours)}} & \tablesize 0.0066 & \tablesize \textbf{0.1466} & \tablesize \textbf{0.2409} & \tablesize \textbf{0.3140} & \tablesize \textbf{0.2732} \\
    \bottomrule
    \end{tabular}
    \egroup
}
\newcommand{\droppedmodetabularkernelfortyktranspose}{
    \bgroup
    \renewcommand{\arraystretch}{0.85}
    \resizebox{\textwidth}{!}{\begin{tabular}{c|c c c c}
        No. Missing & \multirow{2}{*}{Chance} & Improved & \multirow{2}{*}{Coverage} & \textbf{\KGEL} \\
        Modes &   & Recall   &   &   (Ours)   \\
        \midrule
        0 & \textbf{0.0000} & 0.0168 & 0.0012 & 0.0066 \\ %
        2 & 0.3249 & 0.3067 & 0.1700 & \textbf{0.1466} \\ %
        4 & 0.4748 & 0.4363 & 0.3044 & \textbf{0.2409} \\ %
        6 & 0.6063 & 0.5697 & 0.4226 & \textbf{0.3140} \\ %
        8 & 0.7435 & 0.6137 & 0.4805 & \textbf{0.2732} \\ %
    \end{tabular}}
    \egroup
}
\begin{figure*}[t]
    \centering
    \begin{tabular}{ll}
        (a) Mode Dropping & (b) Mode Imbalance \\
         \begin{minipage}{0.45\linewidth}
            \droppedmodecomparisonfortyk
        \end{minipage} &
        \begin{minipage}{0.45\linewidth}
        \centering
            \unbalancedmodecomparisonfortyk
        \end{minipage}
        \\
        \begin{minipage}{0.45\linewidth}
         \droppedmodetabularkernelfortyktranspose
        \end{minipage} &
        \begin{minipage}{0.45\linewidth}
         \unbalancedmodetabularfortyk
        \end{minipage}
        \\
    \end{tabular}
    \vspace{-3mm}
    \caption{\looseness=-1Comparison of Evaluation Metrics on mode dropping (panel (a)), and  mode imbalance (panel (b)). In the mode dropping experiment, up to 8 modes are dropped, and the table on the bottom left is the Hellinger distance between the oracle probability and the calculated one. In the mode imbalance experiment, we calculate the Hellinger distance between oracle and calculated probabilities, and the mode probability changes. %
    }
    \label{fig:unbalanced_mode}
    \vspace{-3mm}
\end{figure*}

%% file: 04_experiments.tex
\section{Experiments}
\label{sec:expts}
\input{figures/validation_experiments/label_corruption}
\subsection{Validation}
\label{ssec:validation}
\textbf{Detecting Mode Dropping} 
\looseness=-1 We repeat the \suman{experiment} in \cref{ssec:motivating_example} with the \textbf{\KGEL{}} test. For our ``generative model'' we use 40,000 images from the CIFAR-10 training set, with 4,000 images per class. We use the remaining 10,000 images in the training set to draw 1,024 witness points. \suman{Otherwise}, the experimental setup is identical.

\looseness=-1 The results in \cref{fig:unbalanced_mode}(a) show improved results of the \KGEL{} test relative to the GEL test and other baselines.
In \cref{ssec:mode_drop_styleGAN}, we also show similar performance when using StyleGAN2+ADA \citep{karras2020training} samples instead of CIFAR-10 training data.

\textbf{Detecting Mode Imbalance} In more realistic scenarios, a generative model may not drop a mode entirely, but instead undersample it relative to other modes. To better understand how accurately the proposed method predicts the degree of mode imbalance, we run a controlled experiment varying the number of examples from two modes. We use the first five categories --- airplane, automobile, bird, cat, and deer --- as the first mode, and the remaining categories as the second. We vary the proportion of examples in the first mode from $0.0$ to $1.0$, and test \KGEL, a normalized improved recall, and coverage to see how well the metrics recover the probabilities. Note that at probabilities $0$ and $1$, this is equivalent to five modes being dropped.

\looseness=-1 The results in Figure \ref{fig:unbalanced_mode}(b) show that our proposed method significantly outperforms competitors. At equal probability, all methods perform similarly. For any degree of unequal probability, \KGEL{} significantly outperforms competitors: the second best method is $29\%$ to $212\%$ worse.

\textbf{Identifying Improper Conditioning with Label Information} When a generative model synthesizes an image from an incorrect class (say a dog instead of butterfly), a test of only images would not identify this issue. This is true of the standard metrics such as FID, IS, and improved precision and recall; and less standard ones such as density coverage.

With \KGEL{} tests that include label information, we can identify the degree of mislabeling. In this test, we take 40,000 images and labels from the CIFAR-10 training set as samples from the generative model distribution. To simulate incorrect conditioning, we change between $10\%$ and $60\%$ of the labels. The CIFAR-10 test set and remaining samples from the training set form the samples from the data distribution and witness points, respectively. The kernel for this test is $k(x, t) = k_d(\phi(x^{(d)}), \phi(t^{(d)})) k_l(x^{(l)}, t^{(l)})$, where $k_d$ is the exponential kernel and $k_l$ the delta kernel. Importantly, in this experiment we assign $\pi_i$ to samples from generative model distribution rather than to those from the data distribution.

\looseness=-1 The results in \cref{fig:label_corruption} %
compare the proposed \KGEL{} test to improved precision and density, both calculated on a per-label basis.\footnote{For this test, we run 10 tests, one for labels airplane, automobile, etc. \suman{\KGEL{}} runs only once.} For improved precision, we vary the value of $k$ to obtain the precision-recall curve, while for density, we vary $k$ to optimize for performance of the area under the precision recall curve.

\looseness=-1 We find that our proposed method significantly outperforms improved precision, and density using $k$ so that expected coverage is greater than $0.95$. One can improve results by artificially increasing $k$. Here we increase it to $100$, over a factor of $30$ of the suggested rate. Even with careful tuning of the baselines, however, \KGEL{} still outperforms these methods without any tuning of its own, suggesting that the proposed method is more \suman{capable of} identifying \suman{improper label conditioning}.

\subsection{Real-World Applications}
\label{ssec:real_world}
\input{figures/least_likely_sample_single_figure/least_likely_biggan_vs_cdm}
\looseness=-1 \textbf{Assessing Within-Class Distributions using Two-Sample Tests}
We use the Kernel Two-sample GEL (\ktsgel) test to better understand how well generative models capture within-class distributions. We use two DGMs in this study --- BigGAN-deep \cite{brock2018large} and the Cascaded Diffusion Model (CDM) \cite{ho2021cascaded}--- as these models represent \suman{among} the best \suman{ examples} of their respective model classes. The \ktsgel{} test uses Pool3 features, 256 witness points from the ImageNet v2 dataset \cite{recht2019imagenet}, and an exponential kernel. The test compares per-class samples from the ImageNet training set and an equivalent number from the DGM. The results in \cref{fig:least_likely_one_figure} show Monarch butterfly samples from the model least representative of the data distribution (model probability $0.0$), and samples from data distribution least representative of the model (data probability $0.0$). BigGAN-deep is least able to represent butterflies that comprise a large portion of the image and swarms of butterflies, while CDMs are least able to represent swarms of butterflies. We show more examples in \cref{ssec:insuff_diversity}.

When we select the 50 samples per class with the lowest model probability (for a total of 50,000 examples), Inception Score of the CDM decreases from 166.2 to 36.82, and of BigGAN-deep from 218.1 to 88.37.

\suman{\looseness=-1\textbf{Understanding Classifier Guidance and the Truncation Trick}}
\input{figures/guidance_scale_vs_truncation/guidance_scale_figure}
DGMs often \suman{employ} a mechanism for trading off sample diversity and sample quality. Two of the most widely used are classifier guidance \cite{sohl2015deep, song2020score} for diffusion models, and the truncation trick \cite{brock2018large} for GANs. In this section, we use the labeled \ktsgel{} test to help us better understand the effect of these methods. We use the Ablated Diffusion Model (ADM) \cite{dhariwal2021diffusion} to measure the effect of classifier guidance, and BigGAN-deep \cite{brock2018large} to measure the effect of the truncation trick.
\cref{fig:guidance_vs_truncation} show the results when the guidance parameter varies from $0.0$ to $10.0$, and the truncation parameter $\tau$ varies from $0.2$ to $1.0$. \suman{As the truncation parameter decreases, unsurprisingly, the number of examples with $0$ data probability increases dramatically
(from $14$ at $\tau=1.0$ to $3420$ at $\tau=0.2$)}. For classifier guidance, we find that optimal performance occurs at scale $1.0$, while increasing the scale beyond this point increases %
the number of examples with $0$ data probability 
to $650$.

\textbf{Model Comparison} %
\label{ssec:model_comparison}
\input{figures/imagenet_model_comparisons/cvpr_main}
We perform a comparison of many popular ImageNet 256$\times$256 models. \cref{tab:EL-imagenet} shows the result for some representative models, while we include a full set of results in \cref{tab:app_EL-imagenet} (we also include results for CIFAR-10 in \cref{ssec:app_c10_model_comparison}). To help delineate between the performance of different models, we report results as $2^{\klkl{\hat{P}_n}{P_{\boldsymbol{\pi}}}}$ and $2^{\klkl{P_{\boldsymbol{\pi}}}{\hat{P}_n}}$ for empirical likelihood, and exponentially tilting GEL, respectively.
We find that performance of BigGAN-deep, ADM with classifier guidance, and VQGAN with $0.05$ acceptance rate perform similarly.%

%% file: figures/validation_experiments/label_corruption.tex
\begin{figure*}
    \centering
           \includegraphics[width=0.32\linewidth]{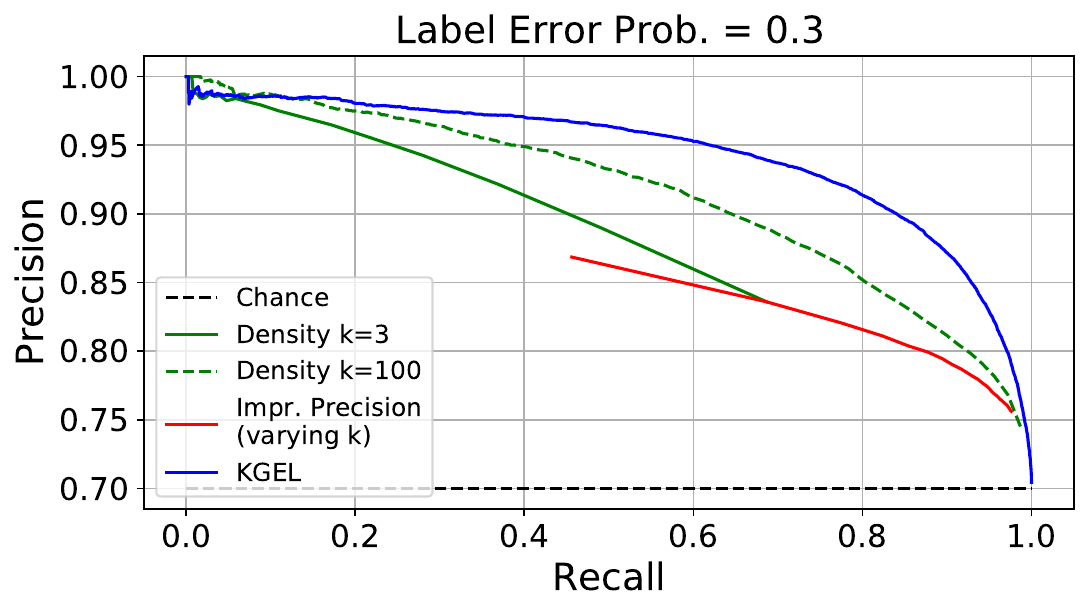} %
     \includegraphics[width=0.32\linewidth]{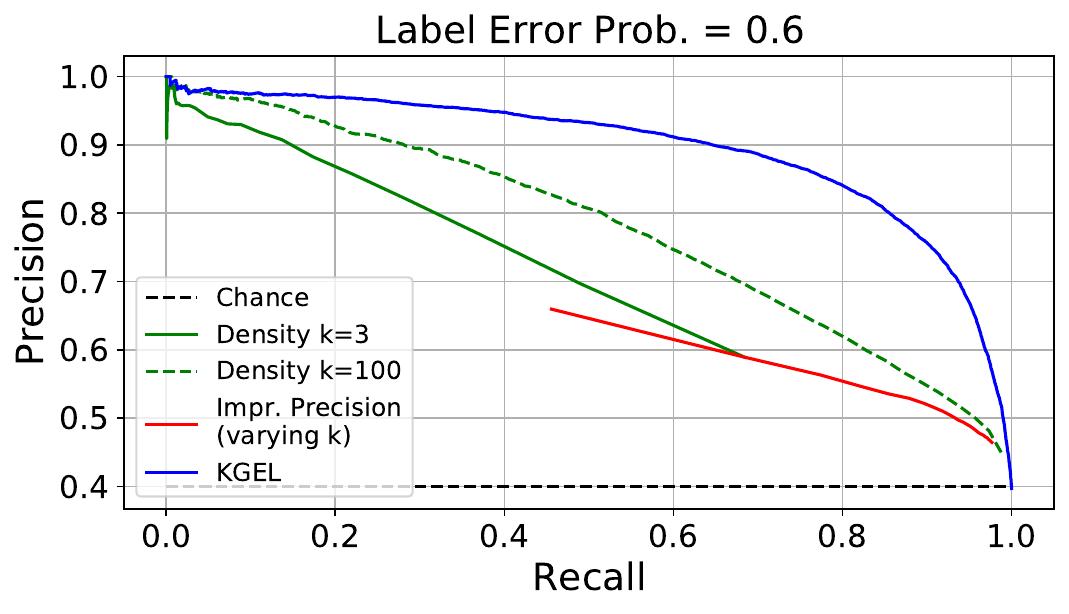} %
     \includegraphics[width=0.34\linewidth]{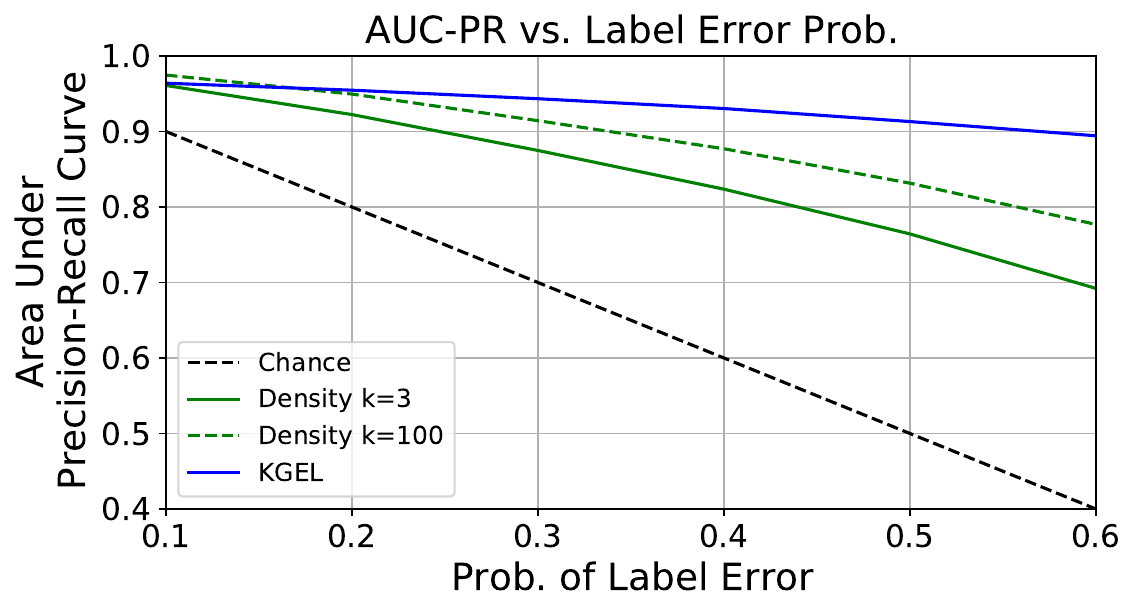} \\
     \vspace{-4mm}
    \caption{Performance of various metrics under label corruption. Left: Precision-Recall Curve at 30\% label error. Middle: Precision-Recall Curve at 60\% label error. Right: Area Under Precision-Recall Curve for different label errors. %
    }
    \vspace{-5mm}
    \label{fig:label_corruption}
\end{figure*}

%% file: figures/least_likely_sample_single_figure/least_likely_biggan_vs_cdm.tex
\begin{figure*}[t]
    \centering
    \setlength{\tabcolsep}{0.5pt}
    \begin{tabular}{c c c}
    \rotatebox[origin=t]{90}{Monarch Butterfly}
    &
    \begin{minipage}{0.49\linewidth}
    \includegraphics[width=0.99\linewidth]{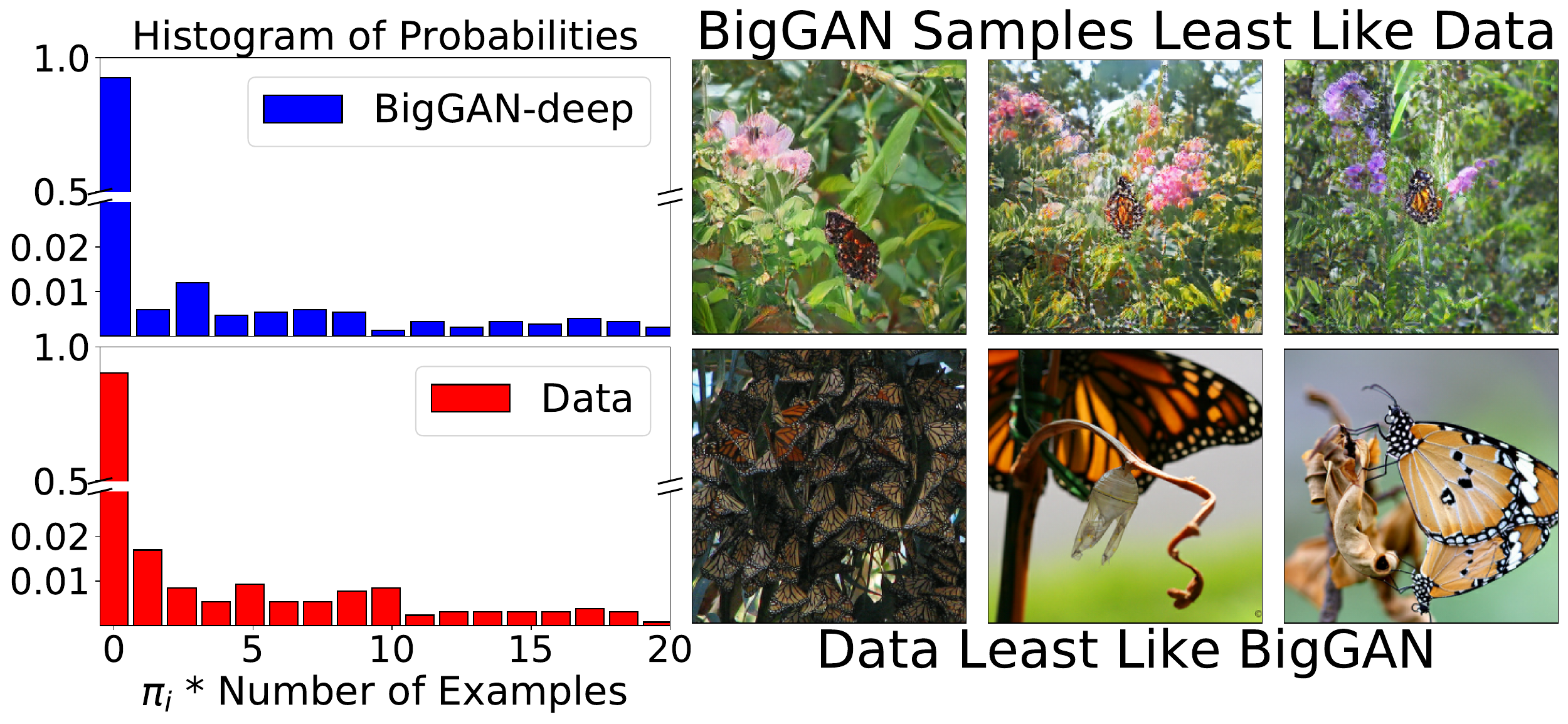} 
    \end{minipage}
    &
    \begin{minipage}{0.49\linewidth}
     \includegraphics[width=0.99\linewidth]{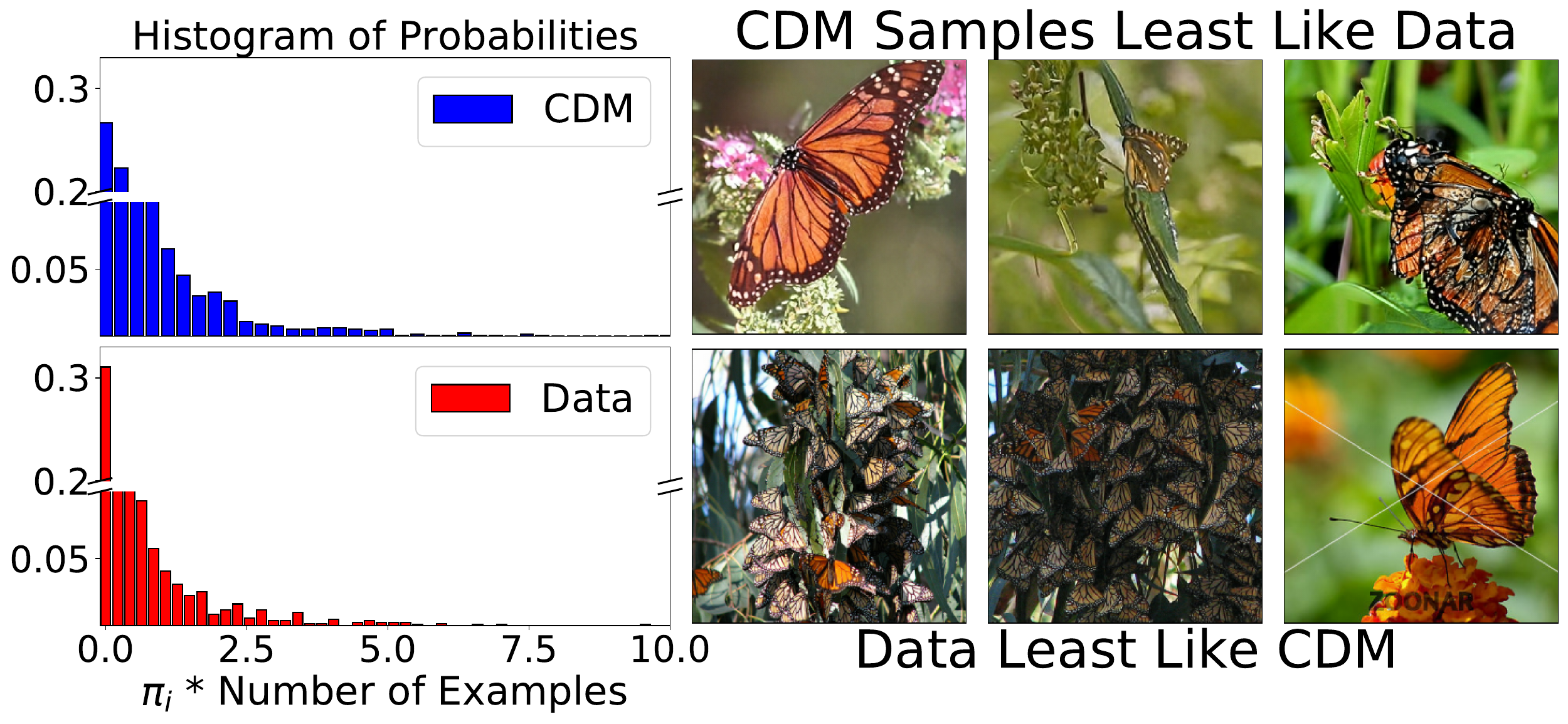}
    \end{minipage}
     \\
    \end{tabular}
    \setlength{\tabcolsep}{6pt}
    \vspace{-3mm}
    \caption{\suman{Model and test probabilities of the \ktsgel{} test can be used to identify data that the model cannot represent, and samples outside the data distribution. We identify examples with $0$ model and data probabilities for BigGAN-deep (left) and Cascaded Diffusion Models (right). The blue and red histograms are those for model and data probabilities, respectively. The three top-right images are model samples least like the data ($0$ model probability), and the bottom-right are examples from the data the model cannot represent ($0$ data probability).}}
    \label{fig:least_likely_one_figure}
\end{figure*}

%% file: figures/guidance_scale_vs_truncation/guidance_scale_figure.tex
\begin{figure*}[t]
\vspace{-3mm}
    \centering
    \begin{tabular}{c c}
        \begin{minipage}{0.25\linewidth}
            \includegraphics[width=\linewidth, trim={0 0 2mm -5mm},clip]{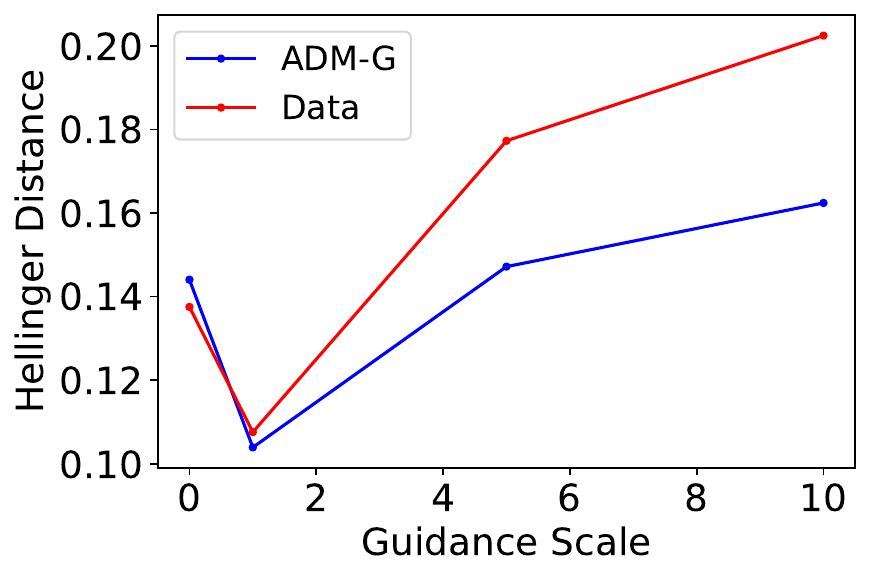}
        \end{minipage}
        &
        \begin{minipage}{0.7\linewidth}
            \includegraphics[width=\linewidth, trim={3mm 0 0 0},clip]{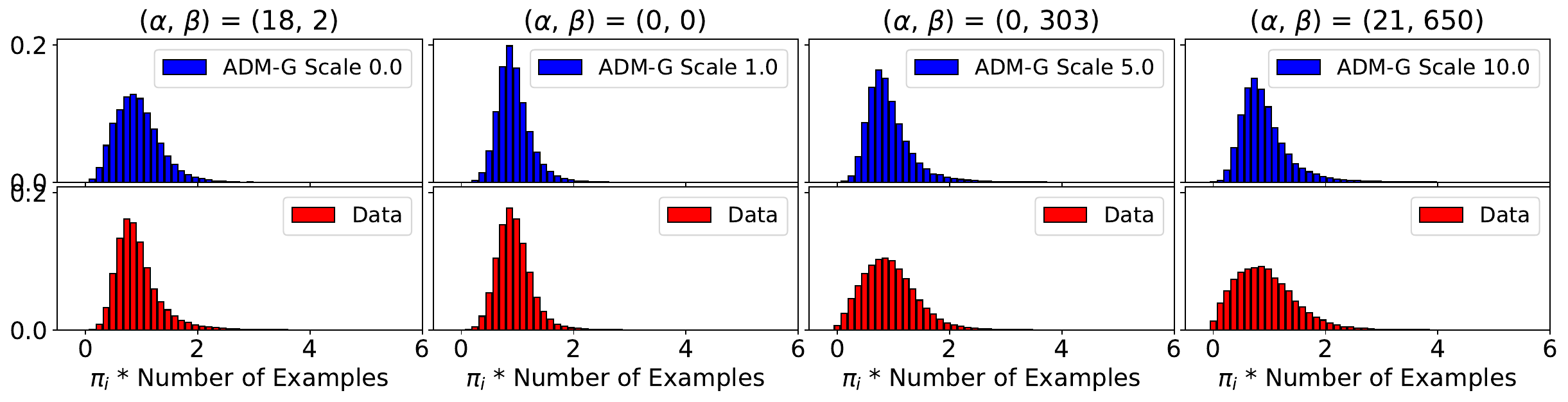}
        \end{minipage}
        \\
        \begin{minipage}{0.25\linewidth}
            \includegraphics[width=\linewidth, trim={0 0 2mm -5mm},clip]{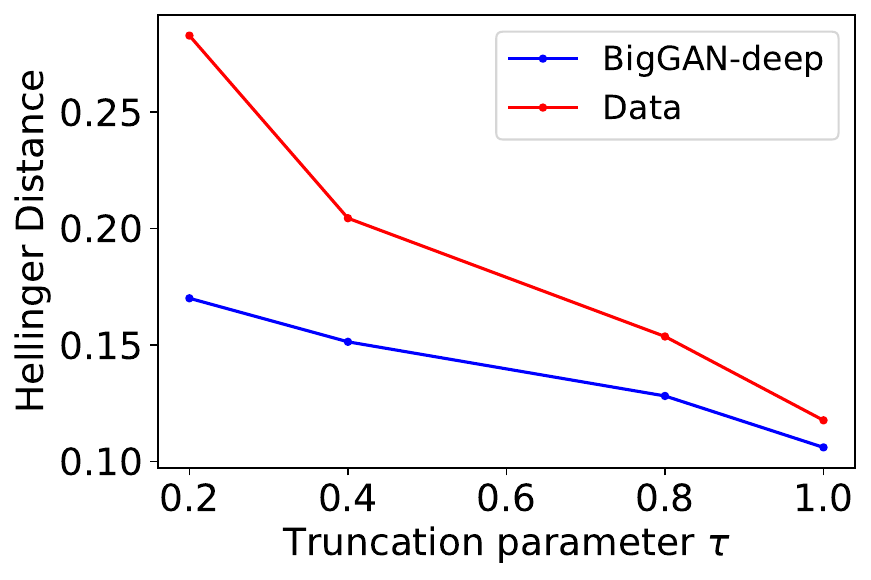}
        \end{minipage}
        &
        \begin{minipage}{0.7\linewidth}
            \includegraphics[width=\linewidth, trim={3mm 0 0 0},clip]{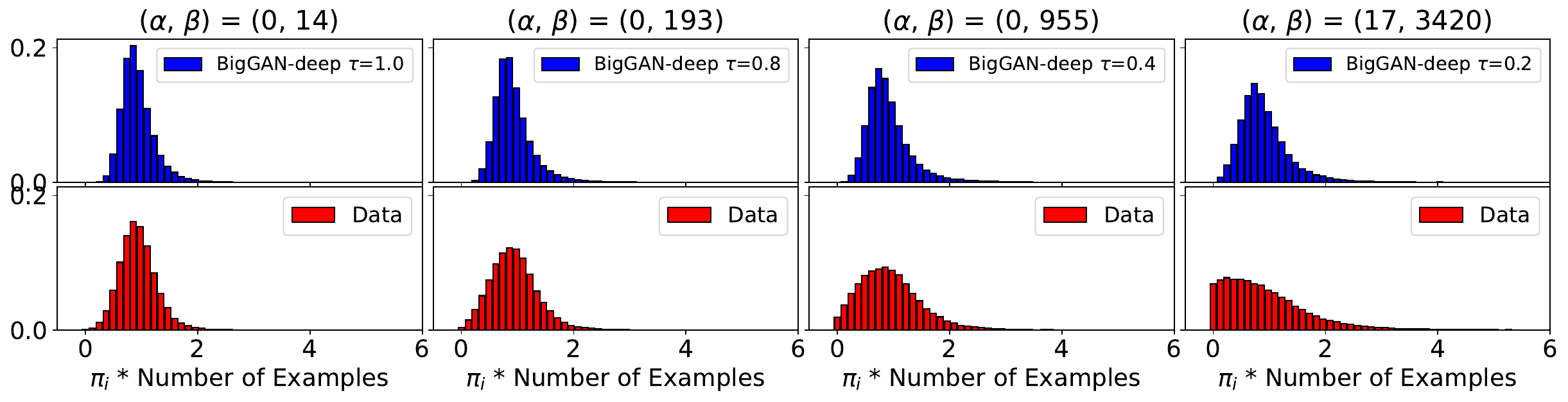} 
        \end{minipage}
        \\
    \end{tabular}
    \vspace{-3mm}
    \caption{Right: Histogram of two-sample probabilities for different ADM classifier Guidance Scales and BigGAN-deep truncation probabilities. We use the labeled \ktsgel{} test with 1000-dim label-balanced ImageNet-v2 witness. \suman{The histograms show how increasing the classifier guidance scale and decreasing the truncation parameter increases the number of examples with $0.0$ data probability. $\alpha$ is the number of DGM samples with $0.0$ model probability, and $\beta$ is the number of examples with $0.0$ data probability. \suman{Left: Hellinger distance between weighted DGM and empirical distributions (blue); and weighted data and empirical distributions (left).}}}
    \vspace{-5mm}
    \label{fig:guidance_vs_truncation}
\end{figure*}

%% file: figures/imagenet_model_comparisons/cvpr_main.tex
\begin{table}[t]
\caption{GEL for Different Models on ImageNet 256$\times$256. Numbers reported as $2^{D(\hat{P}_n\| P_{\boldsymbol{\pi^*}})}$.
For two-sample tests, the numbers reported are model/test \suman{$2^{D(\hat{P}_n\| P_{\boldsymbol{\pi^*}})}$}. Table \ref{tab:app_EL-imagenet} contains a complete set of results. VQGAN$^*$ denotes parameters $k=600$, $t=1.0$, $p=0.92$ and VQGAN$^{**}$ parameters $k=600$, $a=0.05$, $p=1.0$. %
}
\vspace{-4mm}
\label{tab:EL-imagenet}
\begin{center}
\begin{small}
\begin{sc}
\setlength{\tabcolsep}{1pt}
\begin{tabular}{llcccc}
\toprule
\scriptsize \multirow{2}*{Paper} & \scriptsize \multirow{2}*{Model} & \scriptsize \KGEL & \scriptsize \KGEL & \scriptsize \ktsgel & \scriptsize \ktsgel\\
& & \scriptsize EL & \scriptsize ET & \scriptsize EL & \scriptsize ET \\
\midrule
~~- & \scriptsize Theor. opt & \scriptsize 1.0 & \scriptsize 1.0 & \scriptsize 1.0/1.0 & \scriptsize 1.0/1.0\\
\midrule
~~- & \scriptsize Training Set &\scriptsize 1.138 & \scriptsize 1.164 & \scriptsize 1.020/1.018 & \scriptsize 1.021/1.017\\
\midrule
\citet{brock2018large} & \scriptsize BigGAN-deep-$\tau$=1.0 & \scriptsize \textbf{1.735} & \scriptsize \textbf{2.075} & \scriptsize \textbf{1.150}/1.166 & \scriptsize 1.166/1.222 \\
\citet{brock2018large} & \scriptsize BigGAN-deep-$\tau$=0.6 & \scriptsize 2.316 & \scriptsize 3.017 & \scriptsize 1.224/1.271 & \scriptsize  1.260/1.404\\
\midrule
\citet{razavi2019generating}  & \scriptsize VQ-VAE2 & \scriptsize $+\infty$ & \scriptsize 40.55 & \scriptsize 2.933/2.730 & \scriptsize 5.851/5.723 \\
 \citet{esser2021taming} & \scriptsize VQGAN$^*$ & \scriptsize 4.443 & \scriptsize 9.034  & \scriptsize 1.295/1.495 & \scriptsize 1.487/1.717 \\
  \citet{esser2021taming} & \scriptsize VQGAN$^{**}$& \scriptsize 1.772 & \scriptsize 2.219 & \scriptsize 1.175/1.202 & \scriptsize \textbf{1.148}/\textbf{1.158} \\
\midrule
\citet{dhariwal2021diffusion} & \scriptsize ADM & \scriptsize 2.035 & \scriptsize 2.994 & \scriptsize 1.180/1.208 & \scriptsize 1.255/1.244 \\
\citet{dhariwal2021diffusion} & \scriptsize ADM-G (1.0) & \scriptsize 1.786 & \scriptsize 2.289 & \scriptsize 1.155/\textbf{1.151} & \scriptsize 1.185/1.188 \\
\citet{ho2021cascaded} & \scriptsize CDM & \scriptsize 1.857 & \scriptsize 2.467 & \scriptsize 1.161/1.166 & \scriptsize 1.210/1.204\\
\bottomrule
\end{tabular}
\tabsetmedium
\end{sc}
\end{small}
\end{center}
\vskip -0.1in
\vspace{-6mm}
\end{table}

%% file: 05_related_work.tex
\section{Related Work}
\looseness=-1 \suman{In contrast to the amount of research focused on improving generative models, comparatively little is focused on generative model evaluation metrics. There are, however, some notable exceptions. Inception Score (IS) \citep{salimans2016improved} was likely the first broadly used evaluation metric, and measured sample quality and class diversity. %
Followup work such as the Modified Inception Score \citep{Gurumurthy2017} sought to address IS's inability to measure within-class diversity.} %
\looseness=-1 \suman{This class of metrics has largely been replaced by \fid\citep{heusel2017gans}, now the most popular metric for comparing DGMs.} %
\suman{While FID addressed many of the issues of IS, %
it suffered from statistical bias.} %
\suman{Later work \citet{Forsyth2020, binkowski2018demystifying} proposed unbiased alternatives.}

\looseness=-1 \suman{As deep generative models have improved}, a number of authors have proposed more nuanced metrics. A popular approach has been to adopt precision and recall (PR) metrics \citep{sajjadi2018assessing, kynkaanniemi2019improved, ferjad2020reliable, djolonga2020precision}. \suman{These approaches first estimate approximate manifolds of the data and model distributions, and then determine how many data and model samples lie in the model and data manifolds, respectively. The manifold estimation step, however, is highly sensitive to hyperparameters and number of samples.}
\suman{Other ``manifold'' methods include the Geometry Score \citep{khrulkov2018geometry} and the Intrinsic Multi-Scale Distance \citep{Tsitsulin2020}}. %

\looseness=-1 \suman{Metrics that do not fit the above categories include a metric on human evaluation of sample quality \cite{zhou2019hype}, accuracy (on real data) of classifiers trained on DGM data \citep{shmelkov2018good, ravuri2019classification, santurkar2018classification, yang2017lr, esteban2017real}, and for GANs, metrics based on its latent space \citep{Aila2018}.} 

\looseness=-1 %
\suman{Even less literature exists in the machine learning community on Empirical Likelihood methods. Authors proposed an EL test with linear-time MMD constraints \citet{ding2019linear} for testing simple distributions (such as the Normal).} GEL approaches have been proposed for distributionally-robust optimization \citep{duchi2016statistics, lam2017empirical} and for off-policy evaluation \citep{karampatziakis2019empirical, dai2020coindice}. %
Finally, authors introduced lower bounds on GEL objectives with functional moment restrictions for parameter estimation \citep{pmlr-v162-kremer22a}. %

%% file: 06_discussion.tex
\section{Discussion}
\looseness=-1 In this work, we proposed generalized empirical likelihood methods as a tool for a better evaluation of DGMs. We propose a set of interpretable tests that allow us to diagnose deficiencies such as mode dropping and improper label conditioning.
\suman{Current results are promising, and the generality of the approach may lead to new tests. In particular, we are interested in using features of new modalities, such as text, to better evaluate models such as text-conditioned DGMs.}

%% file: 07_appendix.tex
\clearpage
\newpage
\input{07_appendix_further_background}

\input{07_appendix_proofs}

\section{Calculation}
\label{sec:app_calculation}
\suman{\subsection{One-Sample}}
Once we have defined a functions for moment restrictions, we can calculate the the empirical likelihood with relative ease.\footnote{\suman{We follow the derivation in Chapter 3.14 of \cite{owen2001empirical}.}} %
For reference, the original Empirical Likelihood problem with moment constraints is
\begin{align*}
    \begin{aligned}
    &\max_{\{\boldsymbol{\pi} | \sum_i \pi_i=1, \pi_i > 0\}} \sum_{i=1}^n \log{\pi_i} %
    &\mbox{s.t.}\quad
    \E_{\X \sim P_{\pi}}[\vm(\X; \vc)] = \mathbf{0} %
    \end{aligned}
\end{align*}
The Lagrangian is
\begin{align*}
        \mathcal{L}(\pi; \lambda, \nu) = & -\sum_{i=1}^n \log(\pi_i) + \nu \left(\sum_i \pi_i-1\right) \\
        & + \lambda^\top \left(\sum_{i=1}^n \pi_i\vm(x_i; \vc)\right) 
\end{align*}
Solving for $\boldsymbol{\pi}$ and $\nu$ gives us the \suman{dual problem}:
\[ \max_{\lambda} g(\lambda) = \max_{\lambda} \sum_{i=1}^n \log \left(1 + \lambda^\top\vm(x_i; \vc)\right) \]
and $\pi_i^* = \left(n (1 + \lambda^\top\vm(x_i; \vc) \right)^{-1}$. $\pi_i \leq 1$ implies that $1 + \lambda^\top\vm(x_i; \vc) \geq \frac{1}{n}$, which also ensures that $\pi_i \geq 0$. Instead of solving a constrained optimization problem, we modify the logarithm function to be a second order Taylor approximation $1 + \lambda^\top\vm(x_i; \vc) < \frac{1}{n}$ according to \citep{owen2001empirical}.
\[\max_{\lambda} \sum_{i=1}^n \log_{mod} \left(1 + \lambda^\top\vm(x_i; \vc)\right) \]
where $\log_{mod}(z) = \log(1/n) - 1.5 + 2nz - \frac{n^2z^2}{2}$ when $z < \frac{1}{n}$ and the standard logarithm otherwise. For well-defined problems, the optimum is the same. This unconstrained convex objective is easily optimized using Newton's method.

The empirical likelihood is finite if and only if $\boldsymbol{0}$ lies in the interior of the convex hull of $\{\vm(x_i; \vc), \dots \vm(x_n; \vc)\}$ (see  \cref{fig:convex_hull}). If the mean lies outside of the hull, we say that the empirical likelihood is $-\infty$. This requires another algorithm to check if the mean is in the convex hull. We use the triangle algorithm \citep{randomized_triangle}, which tells us that either the point lies outside of the convex hull, or that there exists a point $x$ such that $d(x, \mu) < \epsilon$, where $\epsilon$ is a user-defined parameter. In this latter case, $\mu$ may lie on a boundary point of the convex hull. At the boundary point, there are certain $\pi_i$ that are $0$, also leading to $-\infty$ likelihood. In this case, when solving for $g_m(\lambda)$, $\|\lambda\| \rightarrow \infty$, which is easy to spot during optimization. In practice, we stop optimization when $\|\lambda\| > C$ or $\|\nabla_\lambda g(\lambda)\| > D$, for fixed constants $C, D$. \cref{alg:el} is the pseudocode for the empirical likelihood calculation.

We use the same dual \suman{optimizer} for the exponential tilting objective:  %
\begin{align*}
    &
    \max_{\{\boldsymbol{\pi} | \pi_i \geq 0\}} -\sum_{i=1}^n \pi_i \log{\pi_i} \\
    & 
    \mbox{subject to}\quad
    \E_{\X \sim P_{\pi}}[\vm(\X; \vc)] = \mathbf{0} , 
    \quad
    \sum_{i=1}^n \pi_i = 1%
\end{align*}
The Lagrangian is
\begin{align*}
        \mathcal{L}(\pi; \lambda, \nu) = & \sum_{i=1}^n \pi_i\log(\pi_i) + \nu \left(\sum_i \pi_i-1\right) \\
        & - \lambda^\top \left(\sum_{i=1}^n \pi_i\vm(x_i; \vc)\right).
\end{align*}
Solving for $\boldsymbol{\pi}$ and $\nu$ gives us the \suman{dual problem}:
\begin{align*}
 \max_{\lambda} g(\lambda) &= \max_{\lambda} -\log \left(\sum_{i=1}^n \exp(\lambda^\top\vm(x_i; \vc))\right) \\
 & = \min_{\lambda} \log \left(\sum_{i=1}^n \exp(\lambda^\top\vm(x_i; \vc))\right)
 \end{align*}
with $\pi_i^* = \frac{\exp(\lambda^\top\vm(x_i; \vc))}{\sum_j \exp(\lambda^\top\vm(x_j; \vc))}$. $g(\lambda)$ is concave as the objective is the negative log partition function \cite{wainwright2008graphical}. We compose the log partition function with $\frac{1}{n} \exp(\cdot)$ to give the objective
\[ \min_{\lambda} f(\lambda) = \min_{\lambda} \frac{1}{n}\sum_{i=1}^n \exp(\lambda^\top\vm(x_i; \vc))\]
One can find \suman{an alternative derivation in \cite{kitamura2006empirical}}, and pseudocode for this objective in \cref{alg:et}.
\subsection{Two-Sample}
We derive the dual problem for the two-sample exponential tilting objective for the mean (for general moment restrictions, the calculation is similar). For simplicity, we assume both samples are of the same size. The objective is
\begin{align*}
    &
    \max_{\{\boldsymbol{\pi}, \boldsymbol{\psi} | \pi_i, \psi_j \geq 0\}} -\sum_{i=1}^n \pi_i \log{\pi_i} - \sum_{j=1}^n \psi_j \log{\psi_j}\\
    & 
    \mbox{subject to}\quad
    \sum_{i=1}^n \pi_i x_i = \sum_{j=1}^n \psi_j y_j ,
    \quad
    \sum_{i=1}^n \pi_i = \sum_{j=1}^n \psi_j = 1.%
\end{align*}
Making the following change in variables:
\begin{equation*}
\xi_k \equiv
\begin{cases}
\frac{1}{2} \pi_k, 1 \leq k \leq n\\
\frac{1}{2} \psi_{k-n}, n+1 \leq k \leq 2n \\
\end{cases} 
\end{equation*}
\begin{equation*}
\vz_k \equiv
\begin{cases}
[\vx_k, 1]^{\top}, 1 \leq k \leq n\\
[-\vy_{k-n}, -1]^{\top}, n+1 \leq k \leq 2n \\
\end{cases}
\end{equation*}
We obtain the program:
\begin{align*}
    &
    \max_{\{\boldsymbol{\xi} | \xi_k \geq 0\}} -\sum_{k=1}^{2n} 2\xi_k \log{2\xi_k} \\
    & 
    \mbox{subject to}\quad
    \sum_{k=1}^{2n} \xi_k \vz_k = 0,
    \quad
    \sum_{k=1}^{2n} \xi_k = 1%
\end{align*}
The Lagrangian for this objective (after removing the constant factor of $2$) is
\begin{align*}
    \mathcal{L}(\xi; \lambda, \nu) = & \sum_{k=1}^{2n} \xi_k\log(\xi_k) + \log(2) \sum_{k=1}^{2n} \xi_k
    \\ &+\nu \left(\sum_{k=1}^{2n} \xi_k-1\right) %
    - \lambda^\top \left(\sum_{k=1}^{2n} \xi_k \vz_k \right) 
\end{align*}
We obtain the dual:
\begin{align*}
 \max_{\lambda} g(\lambda) &= \max_{\lambda} -\log \left(\sum_{k=1}^{2n} \exp(\lambda^\top \vz_k) \right) - \log(2) \\
 & = \min_{\lambda} \log \left(\sum_{k=1}^{2n} \exp(\lambda^\top \vz_k )\right) - \log(2)
 \end{align*}

Removing constants and composing with $\frac{1}{n} \exp(\cdot)$ gives us:
\begin{align*}
\min_{\lambda} f(\lambda) = \min_{\lambda} \frac{1}{n} \sum_{k=1}^{2n} \exp(\lambda^\top \vz_k)
\end{align*}
and optimal values:
\begin{align*}
\pi_i^* = 2\frac{\exp(\lambda^\top \vz_i)}{\sum_k \exp(\lambda^\top \vz_k)}, \psi_j^* = 2\frac{\exp(\lambda^\top \vz_{j+n})}{\sum_k \exp(\lambda^\top \vz_k)}
\end{align*}
\input{el_algorithm}
\input{et_algorithm}
\subsection{Calculation Speed}
\suman{As mention in \cref{sec:dgm_tests}, the computational complexity of the method is $\mathcal{O}(nd^3)$. We include further wall clock time results here. \cref{tab:EL-imagenet-timing} shows wall clock time of calculating the metric for ImageNet $256\times256$ models.}

\suman{We also find that performing a full-rank PCA of features prior to solving the GEL objective helps speed up convergence (without changing the objective or results).}

\input{figures/imagenet_model_comparisons/cvpr_timing}
\section{Kernels for Labeled Hierarchies}
When a label hierarchy is available, it provides a more fine-grained way of measuring similarity between labels. 
This is the case for ImageNet labels which are organized in a multi-tree structure with progressively finer categories. Instead of taking into account only the image label, which corresponds to a leaf node, we can use the path from the label to the root of its tree. 
For example, the label ``border collie'' can be represented with the path  \\ organism/animal/verterbrate/mammal/placental/carnivore/ \\
canine/dog/working\_dog/shepherd\_dog/border\_collie, and we can then define a kernel for which ``golden retriever'' will be recognised as more similar to "border collie" than "coffee mug" since they share part of the categories in the path.
We can achieve this using any string kernel to compute the similarity between two paths with nodes acting as ``characters''. One simple option for the kernel is the 
 Smith-Waterman (SW) similarity measure, a type of edit string distance which compares two sequences by calculating the minimum number of transformation operations (e.g.~substitution or gap/deletion) required to convert one sequence into the other. While, in general, SW does not define a valid kernel,  the special case of SW when the substitution matrix score is the identity matrix and there is no penalty for deletions does \cite{vert2004kernel}.    

\section{Further Experimental Results}
\input{figures/imagenet_model_comparisons/cvpr_appendix}
\subsection{Mode Dropping Results with StyleGAN2+ADA}
\label{ssec:mode_drop_styleGAN}
\input{figures/appendix_mode_drop_figure/mode_drop_figure}
We perform the same experiment as in \cref{ssec:validation} using StyleGAN2+ADA \citep{karras2020training} generated samples, which achieved an Inception Score of 10.14 and FID of 2.42 on CIFAR-10. We remove up to 8 classes during sampling (with 5,000 samples per class), similar to the experiment for the training set. We choose $k=3$ for Improved Recall, and $k$ such that expected coverage is $0.95$ (if no modes were dropped) for coverage. We use Pool3 features, and for \KGEL, an exponential kernel with 1,024 witness points from the CIFAR-10 training set. As shown in \cref{fig:app_mode_drops}(a), we find that probabilities estimated using \KGEL{} are sensitive to up to 8 missing classes and outperform baseline methods.

\subsection{CIFAR-10 Model Comparison Results}
\label{ssec:app_c10_model_comparison}
\looseness=-1 For CIFAR-10, we perform a comparison of deep generative models from different model classes to compare performance of both the models and the metric. We perform this comparison across many model classes --- Generative Adversarial Networks \cite{goodfellow2014generative}, Variational Autoencoders \cite{kingma2013auto, rezende2014stochastic}, and diffusion models \cite{sohl2015deep} --- to include a broad range of results. %

\looseness=-1 We perform the \KGEL{} test using the CIFAR-10 test set, 40,000 and 10,000 samples from the model for one-sample and two-sample tests, respectively. We use 1,024 witness points from the CIFAR-10 training set. %
The results in \cref{EL-CIFAR-table} show an important point about the \KGEL{} metric. Only StyleGAN2+ADA and the DDPM have finite score, highlighting how much EL penalizes model misspecification.

These two models stand out as having superior performance compared to over methods across all tests. %
\input{figures/cifar10_model_comparisons/cifar10_appendix}
\subsection{Picking Witness Points}
\label{ssec:witness_features}
\input{figures/appendix_celeba_vs_c10_witness/table}
For \KGEL{} tests, we must choose both what type of dataset we use for our witness points, and how many witness points to use. For the first choice, we find that using a disjoint subset from the same corpus yields the best results. One can, however, use a different corpus and achieve similar results: curiously, we find performance on mode dropping and mode imbalance similar whether we use CIFAR-10 witness points or CelebA \cite{liu2015faceattributes} witness points (\cref{tab:cifar10_vs_celeba}).

For the second, the number of witness points depends on the number of examples in the test set. We found that using roughly 1,000 witness points is a good choice for a test set on the order of 10,000 points. In \cref{fig:app_mode_drops}(b), we repeat the mode dropping experiment in \cref{ssec:validation}, using a varying number of witness points, and find that using 1,024 points yielded the best results.

\section{Further Visualizations}
\subsection{Assessing Within-Class Distributions using Two-Sample Tests}
\label{ssec:insuff_diversity}
\input{figures/least_likely_sample_single_figure/least_likely_biggan_vs_cdm_appendix}
We extend the analysis in \cref{fig:least_likely_one_figure} of \ktsgel{} evaluation of BigGAN-deep and the Cascaded Diffusion Model on per-class ImageNet data and samples. \cref{fig:app_least_likely_one_figure1} and \cref{fig:app_least_likely_one_figure2} show the outputs of \ktsgel{} for different classes on ImageNet.

\input{07_appendix_code}

\newpage
\twocolumn
\section{Theoretical Comparison with Precision-Recall}
 Precision-Recall (PR) and GEL approach the problem of evaluating generated samples from two different perspectives. Sajjadi et al. \cite{sajjadi2018assessing} first define precision and recall for two distributions to quantify how well the generator $q$ “covers” the support of the data distribution $p$ (recall) and how often it generates samples which are unlikely under $p$ (precision). 
 That paper \cite{sajjadi2018assessing} highlights the fact that for samples and data points in the intersection of the supports of $p$ and $q$, there is a fundamental ambiguity: should the difference between $p$ and $q$ be attributed to precision or recall? The authors thus propose to use a continuum of precision-recall values (or precision-recall curve). \cite{kynkaanniemi2019improved} describes important weakness of this approach (ambiguity of using a continuum of value, difficulty to estimate extrema) and argues that the classical definition of PR is sufficient for the task at hand. In practice, we do not have access to $p$ and $q$, only to samples from them, and so to compute PR we first estimate both the data and generated sample manifolds in a feature space. In \cite{kynkaanniemi2019improved} this is done by placing a hypersphere on each point so that it reaches its $k$th nearest neighbor. The resulting PR estimate is especially sensitive to the value of $k$ and to the number of samples used, in particular a larger value of $k$ leads to high values for both precision and recall. 
 
 GEL, on the other hand, is a nonparametrical statistical approach and as such is designed to work directly with samples. In its one-sample version, it attaches a cost to each data point, quantifying how much each point contributes to the mismatch between the data and model distribution. In its two-sided version it also takes into account the symmetric situation, attaching a cost to each model sample. Instead of looking at precision and recall, GEL identifies which samples from the model are not in the data distribution and which data points are not in the model. As such GEL does not require solving the complex intermediate problem of manifold estimation which introduces its own estimation error, hyperparameters, and computational cost. 

%% file: 07_appendix_further_background.tex
\section{Further Empirical Likelihood Background}

\input{figures/single_slide_convex_hull_transpose}

\subsection{Properties of Empirical Likelihood}
\suman{From the perspective of generative model evaluation, the} following are two less important but \suman{nonetheless} interesting properties of the empirical likelihood that, due to space constraints, were not included in the main text.

From a statistical hypothesis testing perspective, if $\E_p[\X] = \vc$, then, similar to the Wilks' Theorem for the Likelihood Ratio Test (LRT) \citep{CaseBerg:01}, the statistic $2n\klkl{\hat{P}_n}{P_{\boldsymbol{\pi}^*}} = -2\sum_{i=1}^n \log(n\pi_i^*)$ converges to a $\chi^2$ distribution.
\begin{theorem}[adapted from \cite{owen2001empirical}]
\label{thm:owen}
Let $x_1,\dots,x_n \in \mathbb{R}^d$ be samples drawn independently from distribution $p$ having mean $\vc$ and finite covariance
matrix $\Sigma$ of rank $q > 0$. Then $-2\sum_{i=1}^n \log(n\pi_i^*) \xrightarrow{d} \chi^2_{(q)}$.
\end{theorem}
Similar results also hold for moment restrictions. Please see Chapter 3.5 of \cite{owen2001empirical}.

We also note that the method has $\mathcal{O}(n^{-1})$ bias, so when one compares models with two-sample tests, one should ensure that the number of model samples should be equal.
\subsection{Convex Hull Condition}
\label{ssec:convex_hull_condition}
The constraints for both empirical likelihood and exponential tilting one-sample tests imply that, for the mean test, objective is finite if and only if the mean can be represented as a convex combination of test points (and for a moment test, $\boldsymbol{0}$ can be expressed as a convex combination of $\vm(x_i; \vc)$). For the empirical likelihood objective, the condition is slightly stronger, as we require $\pi_i > 0$ (otherwise we obtain $\log(0)$). Geometrically, this is equivalent to the mean lying in the interior of the convex hull of test points. For the exponential tilting objective, the mean may also lie on the boundary of the convex hull. For the Euclidean likelihood, the mean may lie anywhere, as the constraint $\pi_i \geq 0$ is removed. The top pane of \cref{fig:convex_hull} shows how the mean may lie within the convex hull for different objectives.

For misspecified DGMs, one or more points may move the mean outside of the convex hull. Furthermore, in high dimensions, the mean may lie on the boundary, or just outside \cite{tsao2004bounds}. For these situations, we can use two-sample methods to extend the utility of GEL methods. In the two-sample case, as shown on the bottom pane of \cref{fig:convex_hull}, we only require the intersection of convex hulls of $\{x_1, \dots, x_n \}$ and $\{y_1, \dots, y_m \}$ to have non-empty interiors for the empirical likelihood objective, and merely non-empty for the exponential tilting objective, for the GEL to provide a finite score. The Euclidean likelihood is always finite.

%% file: figures/single_slide_convex_hull_transpose.tex
\begin{figure}[!t]
\centering
\setlength{\tabcolsep}{3pt}
\begin{tabular}{cccc}
\toprule
& Empirical & Exponential & Euclidean \\
& Likelihood & Tilting & Likelihood \\
\midrule
\rotatebox[origin=c]{90}{\small{One-Sample}} &
\begin{minipage}{0.13\textwidth}
\includegraphics[width=\linewidth, trim=0 0 480 0, clip]{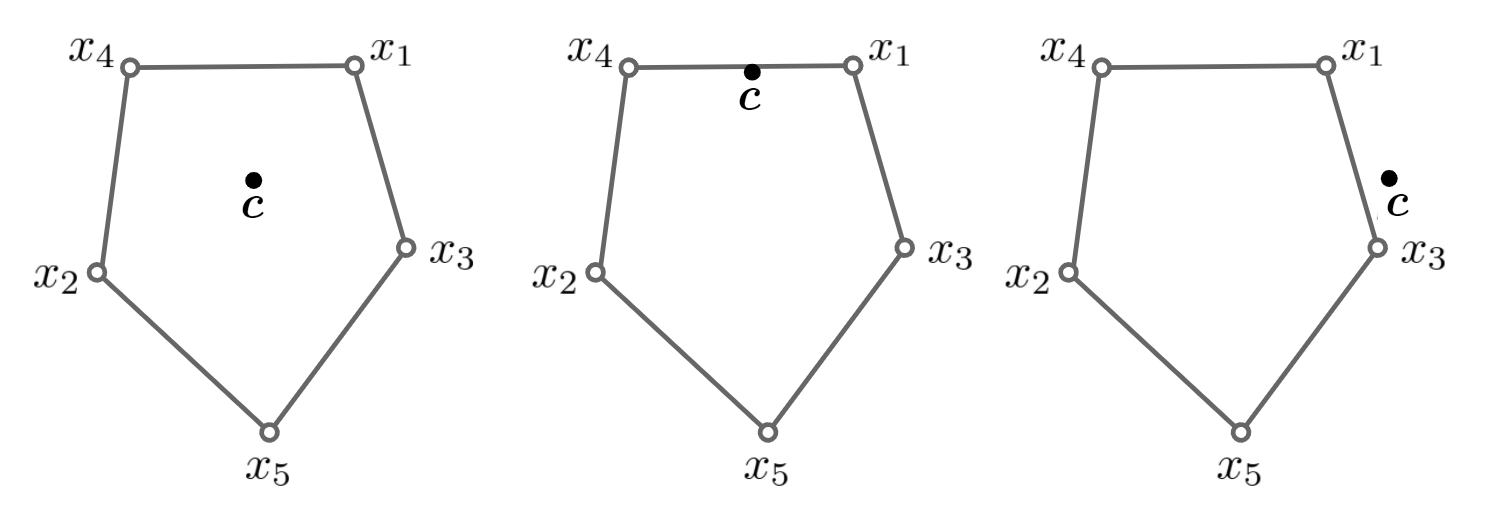}
\end{minipage}
&
\begin{minipage}{0.13\textwidth}
\includegraphics[width=\linewidth, trim=240 0 240 0, clip]{figures/convex_hull/single_slide_convex_hull_v2.png}
\end{minipage}
&
\begin{minipage}{0.13\textwidth}
\includegraphics[width=\linewidth, trim=480 0 0 0, clip]{figures/convex_hull/single_slide_convex_hull_v2.png}
\end{minipage}
\\
\midrule
\rotatebox[origin=c]{90}{\small{Two-Sample}} &
\begin{minipage}{0.14\textwidth}
\includegraphics[width=\linewidth]{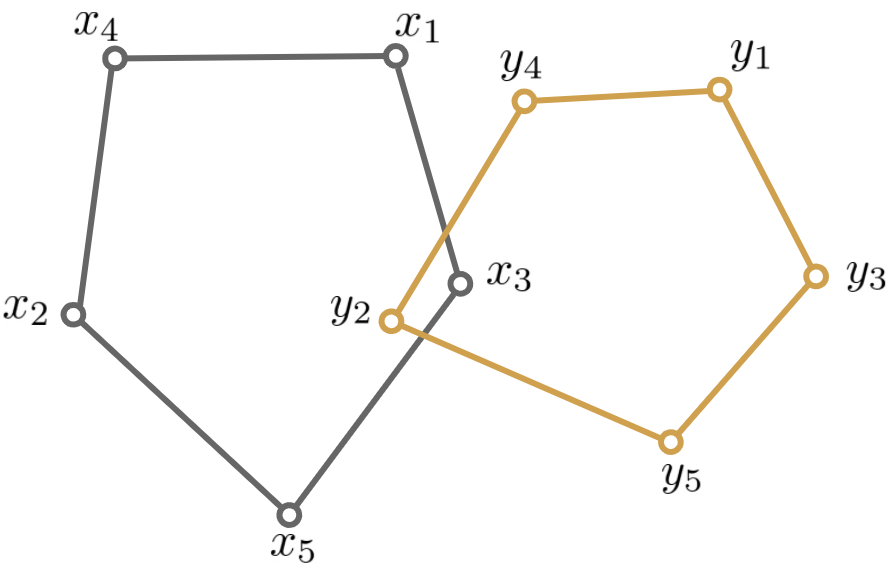}
\end{minipage} &

\begin{minipage}{0.14\textwidth}
\includegraphics[width=\linewidth]{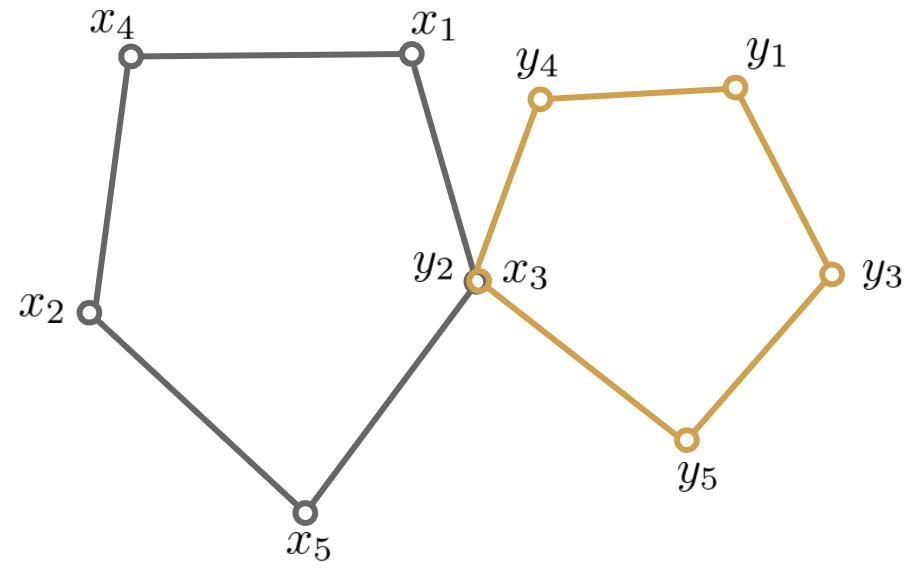}
\end{minipage} &

\begin{minipage}{0.14\textwidth}
\includegraphics[width=\linewidth]{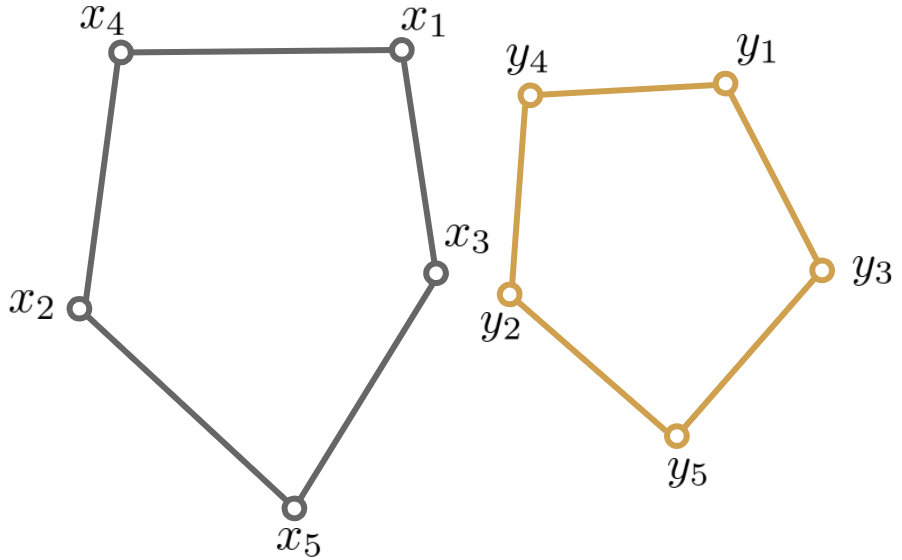}
\end{minipage}
\\
\bottomrule
\end{tabular}
\caption{Illustration of the convex hull condition for one- and two-sample generalized empirical likelihood. For one-sample empirical likelihood to be finite, the $\vc$ vector must lie in the interior of the convex hull (top left). For the exponential tilting to be finite, $\vc$ must lie in the interior or on the boundary of the convex hull (top middle). For two-sample empirical likelihood to be finite, the intersection of the interior of the two convex hulls must be non-empty (bottom left). For two-sample exponential tilting to be finite, the intersection of the closure of the two convex hulls must be non-empty (bottom middle). Euclidean Likelihoods are always finite (top and bottom right).}
\label{fig:convex_hull}
\setlength{\tabcolsep}{6pt}
\end{figure}

%% file: 07_appendix_proofs.tex
\section{Proofs}
\label{sec:app_proofs}

\begin{restatable*}{lemma}{el_orthogonality}
\label{thm:el_orthogonality}
Assume that the true data distribution $p$ is a mixture of the model distribution $q$ and another distribution $\od$. We consider an estimation $\hat{\E}_q [\phi(y)]$ of the model mean obtained using samples $y_i, i \in \mathcal{I} \subset \mathbb{N}$ from $q$; and a test set composed of samples from $p$. The test set can be split into samples $\{\chuck_1, \dots, \chuck_m \}\sim \od$ and samples $\{\keep_{m+1}, \dots, \keep_{n}\}\sim q$. Then, the mean equality condition is 
\begin{align}
\hat{\E}_{q} [\phi(y)] =  \sum_{i=1}^m \pi_i \phi(\chuck_i) + \sum_{i=m+1}^n \pi_i \phi(\keep_i).
\label{eq:el_ort_mean_condition}
\end{align}
If the convex hull $\mathrm{Conv} \{ \phi(\chuck_i), i=1,\dots,m \}$ does not intersect with $ \mathrm{Span} \{ \phi(y_i), i \in \mathcal{I} \cup \{m+1,\dots,n \} \}$ then $\pi_i=0, i=1, \dots, m$.
\end{restatable*}

\begin{proof}
First, note that if $\sum_{i=1}^m \pi_i = 0$ then $\pi_i=0, i=1, \dots, m$ since the weights $\pi_i\geq 0$ are non-negative. Then Lemma~\ref{thm:el_orthogonality} follows. Assume now that $\sum_{i=1}^m \pi_i = a > 0$. Then, we can rewrite the mean condition in~\eqref{eq:el_ort_mean_condition} as
\begin{align}
    \frac{1}{a}\hat{\E}_q [\phi(y)] =  \frac{1}{a}\sum_{i=1}^m \pi_i \phi(\chuck_i) + \frac{1}{a}\sum_{i=m+1}^n \pi_i \phi(\keep_i)\\
    \iff  \frac{1}{a}\hat{\E}_q [\phi(y)] - \frac{1}{a}\sum_{i=m+1}^n \pi_i \phi(\keep_i) = \frac{1}{a}\sum_{i=1}^m \pi_i \phi(\chuck_i).\label{eq:el_ort_rhs}
\end{align}
By moving $a$ into the sum and by writing $a = \sum_{j=1}^m\pi_j$, the right-hand side of \eqref{eq:el_ort_rhs} is given by
\begin{align*}
    &\frac{1}{a}\sum_{i=1}^m \pi_i \phi(\chuck_i) =\sum_{i=1}^m \frac{\pi_i}{\sum_{j=1}^m\pi_j} \phi(\chuck_i)\\
    &\in \mathrm{Conv}\{\phi(\chuck_i), i = 1,\dotsc,m\}
\end{align*}
since $\sum_{i=1}\frac{\pi_i}{\sum_{j=1}^m\pi_j}=1$. For the left-hand side of \eqref{eq:el_ort_rhs} it holds that
\begin{align*}
     &\underbrace{\frac{1}{a}\hat{\E}_q [\phi(y)]}_{\in \mathrm{Span}\{\phi(y_i), i\in\mathcal I\}} - \underbrace{\frac{1}{a}\sum_{i=m+1}^n \pi_i \phi(\keep_i) }_{\in\mathrm{Span}\{\phi(\keep_i), i = m+1,\dots, n\}}\\
     &\in \mathrm{Span}\{\phi(y_i), i \in \mathcal I \cup \{m+1,\dotsc,n\}\}.
\end{align*}
Putting everything together, we obtain a non-trivial intersection of $\mathrm{Span}\{\phi(\keep_i), i \in \mathcal I \cup \{m+1,\dotsc,n\}\}$ and $\mathrm{Conv}\{\phi(\chuck_i), i = 1,\dotsc,m\}$, which contradicts the assumptions of the Lemma. Therefore, $\sum_{i=1}^m\pi_i=0$. 
\end{proof}

%% file: el_algorithm.tex
\begin{algorithm}
\caption{Empirical Likelihood Calculation}
\begin{algorithmic}[1]
\STATE Set $C, D = 1e8, ~ \gamma=1e-8$
\STATE Create $m$ samples from a generative model $y_j \sim q(y)$
\STATE Calculate features $\phi(y_j) \in \mathbb{R}^d$
\STATE Set $\vc = \frac{1}{m}\sum_{i=1}^m \phi(y_j)$
\STATE Calculate features $\phi(x_i) \in \mathbb{R}^d$ for $n$ samples in a test set
\STATE Set $\vm(x_i; \vc) = \phi(x_i) - \vc$
\STATE Check if $\boldsymbol{0} \in Conv\{ \vm(x_i; \vc) \}$ using triangle algorithm \citep{randomized_triangle}
\IF{Convex hull condition fails}
  \STATE return $-\infty$
\ENDIF
\STATE Set $\lambda \in \mathbb{R}^d$ to $\boldsymbol{0}$
\WHILE{$\|\nabla_\lambda g(\lambda)\| > \gamma$ (Not Converged)}
    \STATE Perform Newton Step (see App \ref{ssec:app_el_iteration} for code)
    \IF{$\|\lambda\| > C$ or $\|\nabla_\lambda g(\lambda)\| > D$}
      \STATE return $-\infty$
    \ENDIF
\ENDWHILE
\STATE Set $\pi_i = \frac{1}{n(1 + \lambda^\top \vm(x_i; \vc))}$
\STATE return $\sum_i \log(\pi_i),~~ \boldsymbol{\pi}$
\end{algorithmic}
\label{alg:el}
\end{algorithm}

%% file: et_algorithm.tex
\begin{algorithm}
\caption{Exponential Tilting Calculation}
\begin{algorithmic}[1]
\STATE Set $C, D = 1e8, ~ \gamma=1e-8$
\STATE Create $m$ samples from a generative model $y_j \sim q(y)$
\STATE Calculate features $\phi(y_j) \in \mathbb{R}^d$
\STATE Set $\vc = \frac{1}{m}\sum_{i=1}^m \phi(y_j)$
\STATE Calculate features $\phi(x_i) \in \mathbb{R}^d$ for $n$ samples in a test set
\STATE Set $\vm(x_i; \vc) = \phi(x_i) - \vc$
\STATE Check if $\boldsymbol{0} \in Conv\{ \vm(x_i; \vc) \}$ using triangle algorithm \citep{randomized_triangle}
\IF{Convex hull condition fails}
  \STATE return $-\infty$
\ENDIF
\STATE Set $\lambda \in \mathbb{R}^d$ to $\boldsymbol{0}$
\WHILE{$\|\nabla_\lambda g(\lambda)\| > \gamma$ (Not Converged)}
    \STATE Perform Half-Newton Step (see App \ref{ssec:app_et_iteration} for code)
\ENDWHILE
\STATE Set $\pi_i = \frac{\exp(\lambda^\top \vm(x_i; \vc))}{\sum_j \exp(\lambda^\top \vm(x_i; \vc))}$
\STATE return $-\sum_i \pi_i \log(\pi_i),~~ \boldsymbol{\pi}$
\end{algorithmic}
\label{alg:et}
\end{algorithm}

%% file: figures/imagenet_model_comparisons/cvpr_timing.tex
\begin{table}[t]
\caption{Time (in seconds) needed to calculate GEL for Different Models on ImageNet 256$\times$256. The table does not include time to run the triangle algorithm, or time to calculate or load features. VQGAN$^*$ denotes parameters $k=600$, $t=1.0$, $p=0.92$ and VQGAN$^{**}$ parameters $k=600$, $a=0.05$, $p=1.0$.
}
\vspace{-4mm}
\label{tab:EL-imagenet-timing}
\begin{center}
\begin{small}
\begin{sc}
\setlength{\tabcolsep}{1pt}
\begin{tabular}{llcccc}
\toprule
\scriptsize \multirow{2}*{Paper} & \scriptsize \multirow{2}*{Model} & \scriptsize \KGEL & \scriptsize \KGEL & \scriptsize \ktsgel & \scriptsize \ktsgel\\
& & \scriptsize EL & \scriptsize ET & \scriptsize EL & \scriptsize ET \\
\midrule
~~- & \scriptsize Training Set &\scriptsize 29.96 & \scriptsize 112.3 & \scriptsize 28.26 & \scriptsize 146.1 \\
\midrule
\citet{brock2018large} & \scriptsize BigGAN-deep-$\tau$=1.0 & \scriptsize 133.9 & \scriptsize 113.4 & \scriptsize 67.99 & \scriptsize 188.7 \\
\citet{brock2018large} & \scriptsize BigGAN-deep-$\tau$=0.6 & \scriptsize 215.2 & \scriptsize 81.53 & \scriptsize 62.13 & \scriptsize 177.7  \\
\midrule
\citet{razavi2019generating}  & \scriptsize VQ-VAE2 & \scriptsize 24.93 & \scriptsize 92.68 & \scriptsize 243.9 & \scriptsize 211.7 \\
 \citet{esser2021taming} & \scriptsize VQGAN$^*$ & \scriptsize 118.14 & \scriptsize 72.50 & \scriptsize 138.3 & \scriptsize 179.9 \\
  \citet{esser2021taming} & \scriptsize VQGAN$^{**}$& \scriptsize 76.83 & \scriptsize 67.85 & \scriptsize 68.3 & \scriptsize 173.3 \\
\midrule
\citet{dhariwal2021diffusion} & \scriptsize ADM & \scriptsize 129.4 & \scriptsize 80.50 & \scriptsize 75.39 & \scriptsize 154.4 \\
\citet{dhariwal2021diffusion} & \scriptsize ADM-G (1.0) & \scriptsize 120.4 & \scriptsize 75.37 & \scriptsize 64.31 & \scriptsize 187.4 \\
\citet{ho2021cascaded} & \scriptsize CDM & \scriptsize 83.25 & \scriptsize 64.24 & \scriptsize 67.4 & \scriptsize 188.1 \\
\bottomrule
\end{tabular}
\tabsetmedium
\end{sc}
\end{small}
\end{center}
\vskip -0.1in
\end{table}

%% file: figures/imagenet_model_comparisons/cvpr_appendix.tex
\begin{table}[t]
\caption{\looseness=-1 Expanded results of Table \ref{tab:EL-imagenet}: GEL for Different Models on ImageNet 256$\times$256. Numbers are reported as $2^{D(\hat{P}_n\| P_{\boldsymbol{\pi^*}})}$, so as to better delineate performance among models. For two-sample tests, the numbers reported are model/test \suman{$2^{D(\hat{P}_n\| P_{\boldsymbol{\pi^*}})}$}. VQGAN$^*$ denotes parameters $k=600$, $t=1.0$, $p=0.92$ and VQGAN$^{**}$ parameters $k=600$, $a=0.05$, $p=1.0$. For BigGAN-deep, ``T'' denotes the truncation parameter. For ADM-G, the number in parenthesis denotes the guidance scale. ``-'' indicated the optimizer failed. %
}
\label{tab:app_EL-imagenet}
\begin{center}
\begin{small}
\begin{sc}
\setlength{\tabcolsep}{1pt}
\begin{tabular}{llcccc}
\toprule
\scriptsize \multirow{2}*{Paper} & \scriptsize \multirow{2}*{Model} & \scriptsize \KGEL & \scriptsize \KGEL & \scriptsize \ktsgel & \scriptsize \ktsgel\\
& & \scriptsize EL & \scriptsize ET & \scriptsize EL & \scriptsize ET \\
\midrule
~~- & \scriptsize Theor. opt & \scriptsize 1.0 & \scriptsize 1.0 & \scriptsize 1.0 & \scriptsize 1.0\\
\midrule
~~- & \scriptsize Training Set &\scriptsize 1.138 & \scriptsize 1.164 & \scriptsize 1.020/1.018 & \scriptsize 1.021/1.017\\
\midrule
\citet{brock2018large} & \scriptsize BigGAN-deep-t=1.0 & \scriptsize \textbf{1.735} & \scriptsize \textbf{2.075} & \scriptsize \textbf{1.150}/1.166 & \scriptsize 1.166/1.222 \\
\citet{brock2018large} & \scriptsize BigGAN-deep-t=0.8 & \scriptsize 1.964 & \scriptsize 2.443 & \scriptsize 1.183/1.201 & \scriptsize  1.203/1.288\\
\citet{brock2018large} & \scriptsize BigGAN-deep-t=0.6 & \scriptsize 2.316 & \scriptsize 3.017 & \scriptsize 1.224/1.271 & \scriptsize  1.260/1.404\\
\citet{brock2018large} & \scriptsize BigGAN-deep-t=0.2 & \scriptsize 3.965 & \scriptsize 6.354 & \scriptsize -/- & \scriptsize  1.396/1.955\\
\midrule
\citet{razavi2019generating}  & \scriptsize VQ-VAE2 & \scriptsize $+\infty$ & \scriptsize 40.55 & \scriptsize 2.933/2.730 & \scriptsize 5.851/5.723 \\
 \citet{esser2021taming} & \scriptsize VQGAN$^*$ & \scriptsize 4.443 & \scriptsize 9.034  & \scriptsize 1.295/1.495 & \scriptsize 1.487/1.717 \\
  \citet{esser2021taming} & \scriptsize VQGAN$^{**}$& \scriptsize 1.772 & \scriptsize 2.219 & \scriptsize 1.175/1.202 & \scriptsize \textbf{1.148}/\textbf{1.158} \\
\midrule
\citet{dhariwal2021diffusion} & \scriptsize ADM & \scriptsize 2.035 & \scriptsize 2.994 & \scriptsize 1.180/1.208 & \scriptsize 1.255/1.244 \\
\citet{dhariwal2021diffusion} & \scriptsize ADM-G (1.0) & \scriptsize 1.786 & \scriptsize 2.289 & \scriptsize 1.155/\textbf{1.151} & \scriptsize 1.185/1.188 \\
\citet{dhariwal2021diffusion} & \scriptsize ADM-G (5.0) & \scriptsize 2.492 & \scriptsize 3.602 & \scriptsize 1.232/1.232 & \scriptsize 1.294/1.324 \\
\citet{dhariwal2021diffusion} & \scriptsize ADM-G (10.0) & \scriptsize 3.219 & \scriptsize 5.160 & \scriptsize 1.279/1.299 & \scriptsize 1.378/1.442 \\
\citet{ho2021cascaded} & \scriptsize CDM & \scriptsize 1.857 & \scriptsize 2.467 & \scriptsize 1.161/1.166 & \scriptsize 1.210/1.204\\
\bottomrule
\end{tabular}
\tabsetmedium
\end{sc}
\end{small}
\end{center}
\vskip -0.1in
\end{table}

%% file: figures/appendix_mode_drop_figure/mode_drop_figure.tex
\newcommand{\droppedmodetabularkernelfortyktransposestylegan}{
    \bgroup
    \renewcommand{\arraystretch}{0.85}
    \resizebox{\textwidth}{!}{\begin{tabular}{c|c c c c}
        No. Missing & \multirow{2}{*}{Chance} & Improved & \multirow{2}{*}{Coverage} & \textbf{\KGEL} \\
        Modes &   & Recall   &   &   (Ours)   \\
        \midrule
        0 & \textbf{0.0000} & 0.0281 & 0.0059 & 0.0176 \\
        2 & 0.3249 & 0.3317 & 0.1643 & \textbf{0.1413} \\
        4 & 0.4748 & 0.4723 & 0.3180 & \textbf{0.2503} \\
        6 & 0.6063 & 0.6128 & 0.4211 & \textbf{0.3326} \\
        8 & 0.7435 & 0.6723 & 0.4820 & \textbf{0.3405} \\
    \end{tabular}}
    \egroup
}
\newcommand{\droppedmodetabularkernelnumwitnesspoints}{
    \bgroup
    \renewcommand{\arraystretch}{0.85}
    \resizebox{\textwidth}{!}{\begin{tabular}{c|c c c c}
        No. Missing & \multirow{2}{*}{64-dim} & \multirow{2}{*}{256-dim} & \multirow{2}{*}{1024-dim} & \multirow{2}{*}{2048-dim} \\
        Modes &   &    &   &   \\
        \midrule
        0 & 0.0055 & \textbf{0.0051} & 0.0066 & 0.0117 \\
        2 & 0.1973 & 0.1679 & \textbf{0.1466} & 0.2164 \\
        4 & 0.3193 & 0.2704 & \textbf{0.2409} & 0.3064 \\
        6 & 0.3890 & 0.3485 & \textbf{0.3140} & 0.3905 \\
        8 & 0.3584 & 0.3208 & \textbf{0.2732} & 0.4294 \\
    \end{tabular}}
    \egroup
}

\newcommand{\stylegandroppedmodecomparisonfortyk}{
    \includegraphics[width=\linewidth, trim=0 0 0 0, clip]{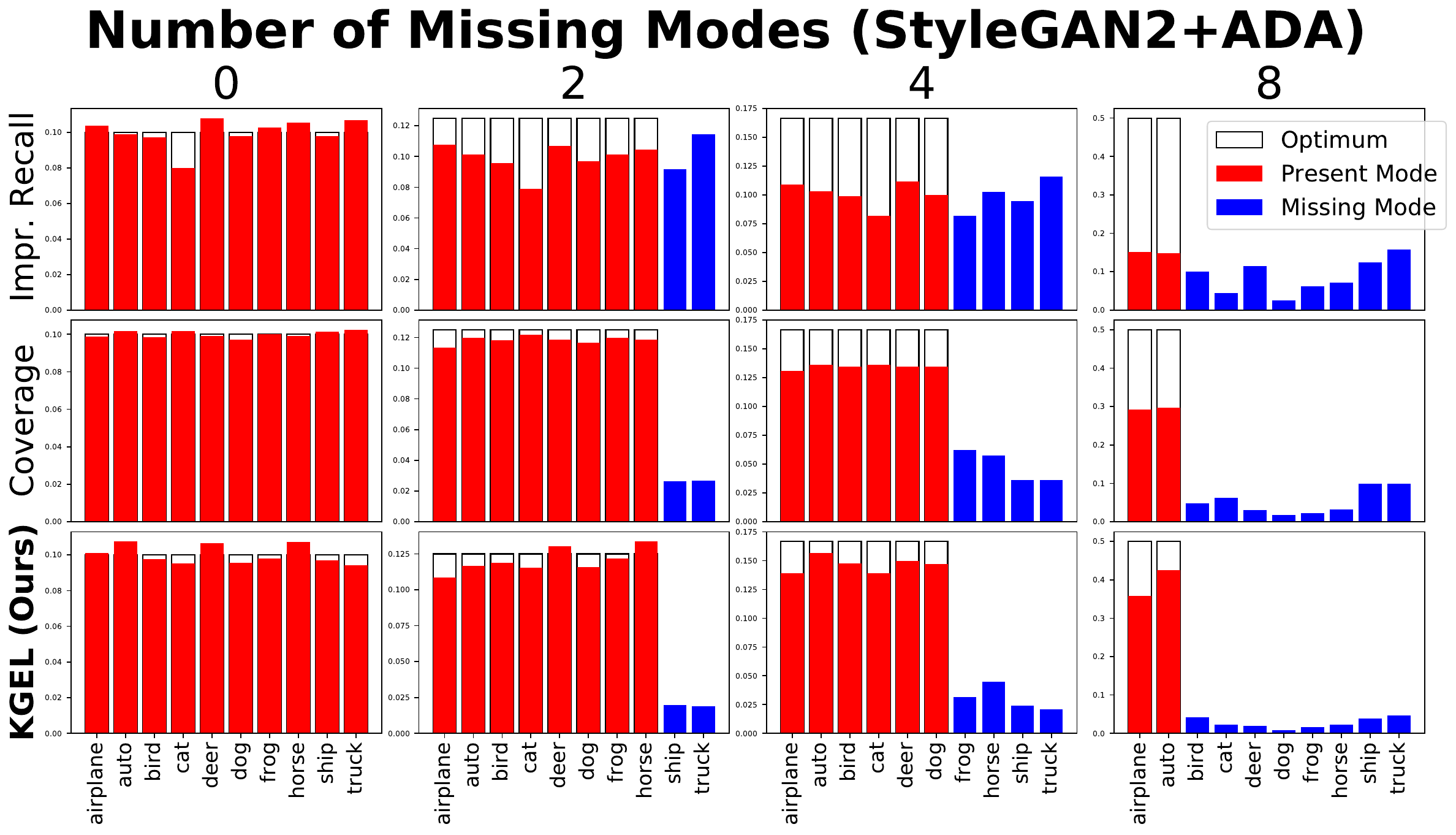}
    }
    
\newcommand{\witnessdroppedmodecomparisonfortyk}{
    \includegraphics[width=\linewidth, trim=0 0 0 0, clip]{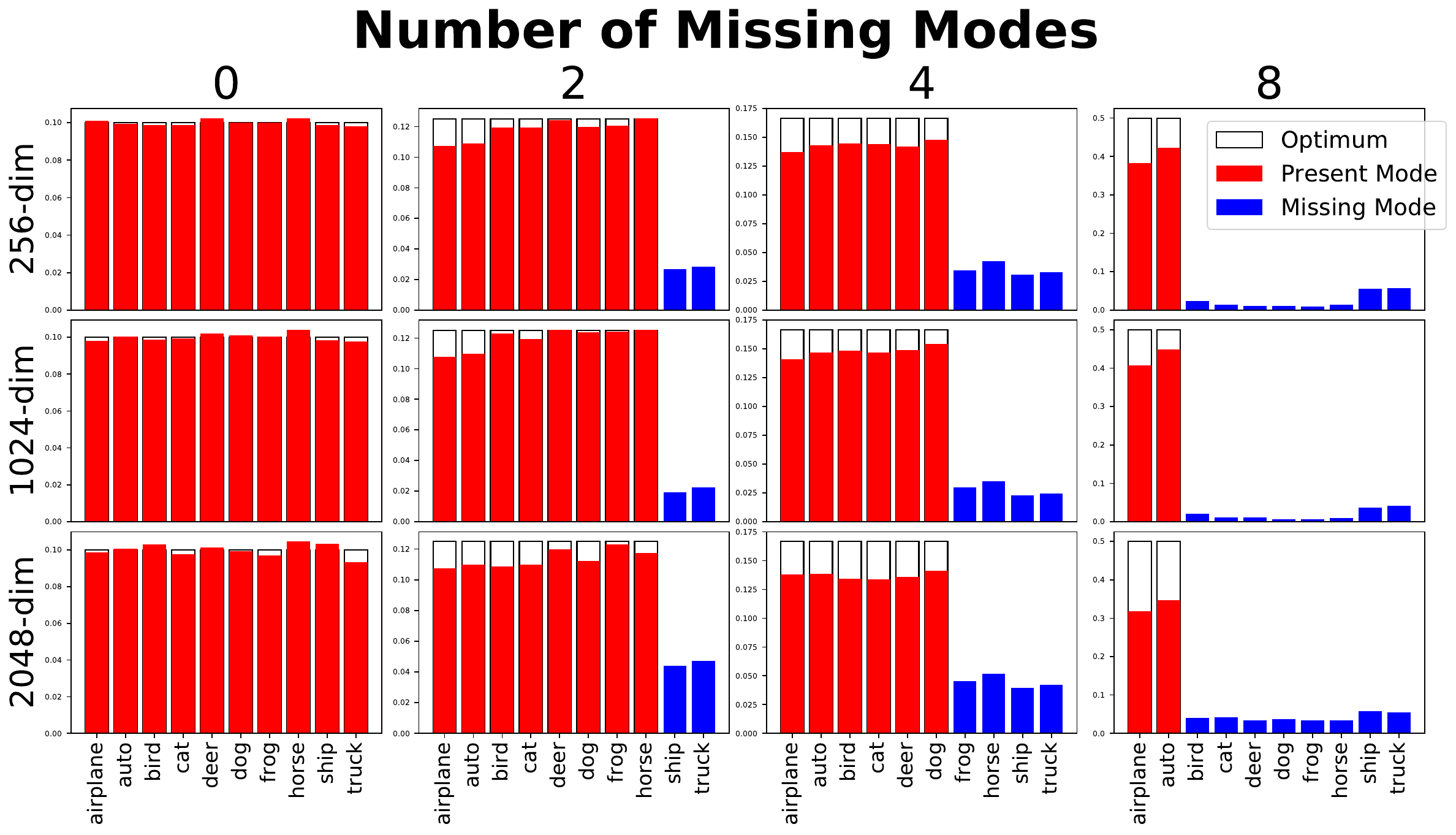}
    }

\begin{figure*}[t]
    \centering
    \begin{tabular}{ll}
        (a) Mode Dropping for StyleGAN2+ADA & (b) \KGEL{} Mode Dropping by No. of Witness Points \\
         \begin{minipage}{0.45\linewidth}
            \stylegandroppedmodecomparisonfortyk
        \end{minipage} &
        \begin{minipage}{0.45\linewidth}
        \centering
            \witnessdroppedmodecomparisonfortyk
        \end{minipage}
        \\
        \begin{minipage}{0.45\linewidth}
         \droppedmodetabularkernelfortyktransposestylegan
        \end{minipage} &
        \begin{minipage}{0.45\linewidth}
         \droppedmodetabularkernelnumwitnesspoints
        \end{minipage}
        \\
    \end{tabular}
    \vspace{-3mm}
    \caption{\looseness=-1\textbf{(a)} Comparison of Evaluation Metrics on mode dropping of StyleGAN2+ADA classes, and \textbf{(b)} evaluation of performance on CIFAR-10 mode dropping of the \KGEL{} tests based on the number of witness points. In both experiments, up to 8 modes are dropped, and the tables on the bottom are the Hellinger distance between the oracle probability and the calculated one.
    }
    \label{fig:app_mode_drops}
\end{figure*}

%% file: figures/cifar10_model_comparisons/cifar10_appendix.tex
\begin{table}[ht]
\setlength{\tabcolsep}{3pt}
\caption{GEL for Different Models on CIFAR-10. Numbers are reported as $2^{D(\hat{P}_n\| P_{\boldsymbol{\pi^*}})}$, so as to better delineate performance among models.}
\label{EL-CIFAR-table}
\scriptsize
\begin{center}
\begin{sc}
\begin{tabular}{cllcccc}
\toprule
& \multirow{2}*{Paper} & \multirow{2}*{Model} & \KGEL & \KGEL & \ktsgel & \ktsgel\\
& & & EL & ET & EL & ET \\
\midrule
& ~~- & Theor. Opt.	& 1.0 & 1.0 &  1.0/1.0 & 1.0/1.0 \\
& ~~- & \scriptsize Training Set & 1.198 & 1.217 & 1.026/1.027 & 1.027/1.027 \\
\midrule
\multirow{4}{*}{\rotatebox[origin=c]{90}{\scriptsize{VAE}}} 
& \citep{razavi2019preventing} & Delta VAE &$+\infty$ & 12.037 & 2.088/2.125 & 2.506/2.622 \\
& \citep{vahdat2020nvae} & NVAE-$\tau=$0.7 & $+\infty$ & 11.078 & 2.556/2.666 & 3.158/3.513\\
& \citep{vahdat2020nvae} & NVAE-$\tau=$1.0 & $+\infty$ & 11.644 & 2.108/2.266 & 2.490/2.908\\
& \citep{child2020very} & VD-VAE & $+\infty$ & 11.644 & 2.560/2.891 & 3.171/4.049\\
\midrule
\multirow{5}{*}{\rotatebox[origin=c]{90}{{GAN}}} 
& \citep{miyato2018spectral} & SNGAN & $+\infty$& 13.454 & 1.584/1.730 & 1.740/2.030\\
&  \citep{ravuri2018learning} & MoLM-1024 &$+\infty$ & 12.542 & 1.649/1.751 &1.856/2.047 \\
& \cite{karras2017progressive} & ProGAN & $+\infty$& 13.765 & 1.414/1.528 & 1.510/1.675\\
& \citep{brock2018large} & BigGAN &$+\infty$ & 12.667 & 1.549/1.613 & 1.708/1.819\\
&  \citep{karras2020training} & StyleGAN2+ADA & \textbf{1.632} & \textbf{1.724} & \textbf{1.094/1.092} & \textbf{1.101/1.100}\\
\midrule
\multirow{4}{*}{\rotatebox[origin=c]{90}{\tiny{Score-Based}}} 
& \citep{song2019generative} & NCSN-v1 &$+\infty$ & 10.670 & 2.312/2.147 & 2.855/2.586 \\
& \citep{song2020improved} & NCSN-v2 & $+\infty$& 9.324& 3.042/2.679 & 3.915/3.193\\
&  \citep{song2020improved} & NCSN-v2 (w/denoi) & $+\infty$& 13.830& 1.346/1.388 & 1.426/1.482\\
& \citep{ho2020denoising} & DDPM & 1.716 & 1.804 & 1.108/1.113 & 1.113/1.125\\
\bottomrule
\end{tabular}
\end{sc}
\end{center}
\vskip -0.1in
\setlength{\tabcolsep}{6pt}
\end{table}

%% file: figures/appendix_celeba_vs_c10_witness/table.tex
\newcommand{\droppedmodesbywitnesstype}{
    \bgroup
    \renewcommand{\arraystretch}{0.85}
    \resizebox{0.45\linewidth}{!}{\begin{tabular}{c| c c}
        \toprule
        No. Missing & CIFAR-10 & CelebA  \\
        Modes & Witness & Witness  \\
        \midrule
        0 & 0.0066 & 0.0074 \\
        2 & 0.1466 & 0.1548 \\
        4 & 0.2409 & 0.2503 \\
        6 & 0.3140 & 0.3226 \\
        8 & 0.2732 & 0.2861 \\
        \bottomrule
    \end{tabular}}
    \egroup
}
\newcommand{\unbalancedmodebywitnesstype}{
    \bgroup
    \renewcommand{\arraystretch}{0.85}
    \resizebox{0.43\linewidth}{!}{\begin{tabular}{c| c c}
        \toprule
        Mode 1 & CIFAR-10 & CelebA  \\
        Probability & Witness & Witness  \\
        \midrule
          0.1 & 0.1206 & 0.1258 \\
         0.3 & 0.0486 & 0.0508 \\
         0.5 &  0.0030 & 0.0029 \\
         0.7 & 0.0434 &  0.0475\\
         0.9 & 0.1207 & 0.1295 \\
        \bottomrule
    \end{tabular}}
    \egroup
}
\begin{table}[ht]
    \caption{Comparison of performance of \KGEL{} tests on Mode Dropping (left) and Mode Imbalance (right) using 1,024 CIFAR-10 and CelebA Witness Points.}
    \centering
    \begin{tabular}{c c}
        Mode Dropping & Mode Imbalance \\
        \droppedmodesbywitnesstype & \unbalancedmodebywitnesstype \\
    \end{tabular}
    \label{tab:cifar10_vs_celeba}
\end{table}

%% file: figures/least_likely_sample_single_figure/least_likely_biggan_vs_cdm_appendix.tex
\newcommand{\leastlikelysinglefigure}[2]{
    \rotatebox[origin=t]{90}{#1}
    &
    \begin{minipage}{0.49\linewidth}
    \includegraphics[width=0.99\linewidth]{figures/least_likely_sample_single_figure/imagenet_v2_witness/biggan_vs_real_probs_broken_bar_imagenet_v2_class#2_v3.pdf} 
    \end{minipage}
    &
    \begin{minipage}{0.49\linewidth}
     \includegraphics[width=0.99\linewidth]{figures/least_likely_sample_single_figure/imagenet_v2_witness/cdm_vs_real_probs_broken_bar_imagenet_v2_class#2_v3.pdf}
    \end{minipage}
}

\newcommand{\leastlikelysinglefigurelowerres}[2]{
    \rotatebox[origin=t]{90}{#1}
    &
    \begin{minipage}{0.49\linewidth}
    \includegraphics[width=0.99\linewidth]{figures/least_likely_sample_single_figure/imagenet_v2_witness/biggan_vs_real_probs_broken_bar_imagenet_v2_class#2_v3_lower_res.pdf} 
    \end{minipage}
    &
    \begin{minipage}{0.49\linewidth}
     \includegraphics[width=0.99\linewidth]{figures/least_likely_sample_single_figure/imagenet_v2_witness/cdm_vs_real_probs_broken_bar_imagenet_v2_class#2_v3_lower_res.pdf}
    \end{minipage}
}

\begin{figure*}[t]
    \centering
    \setlength{\tabcolsep}{0.5pt}
    \begin{tabular}{c c c}
    \leastlikelysinglefigure{Monarch Butterfly}{323} \\
    \leastlikelysinglefigure{Jackfruit}{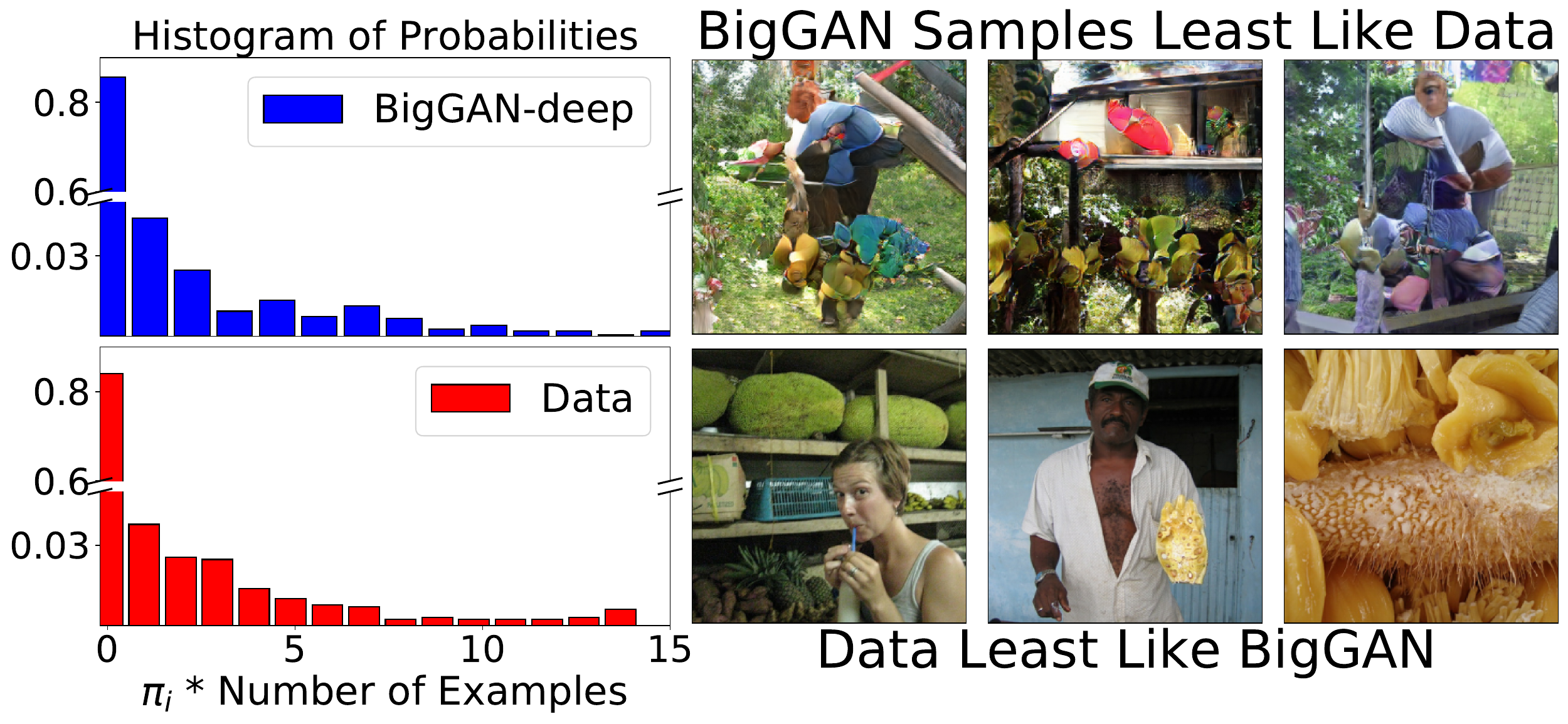} \\
    \leastlikelysinglefigure{Stethoscope}{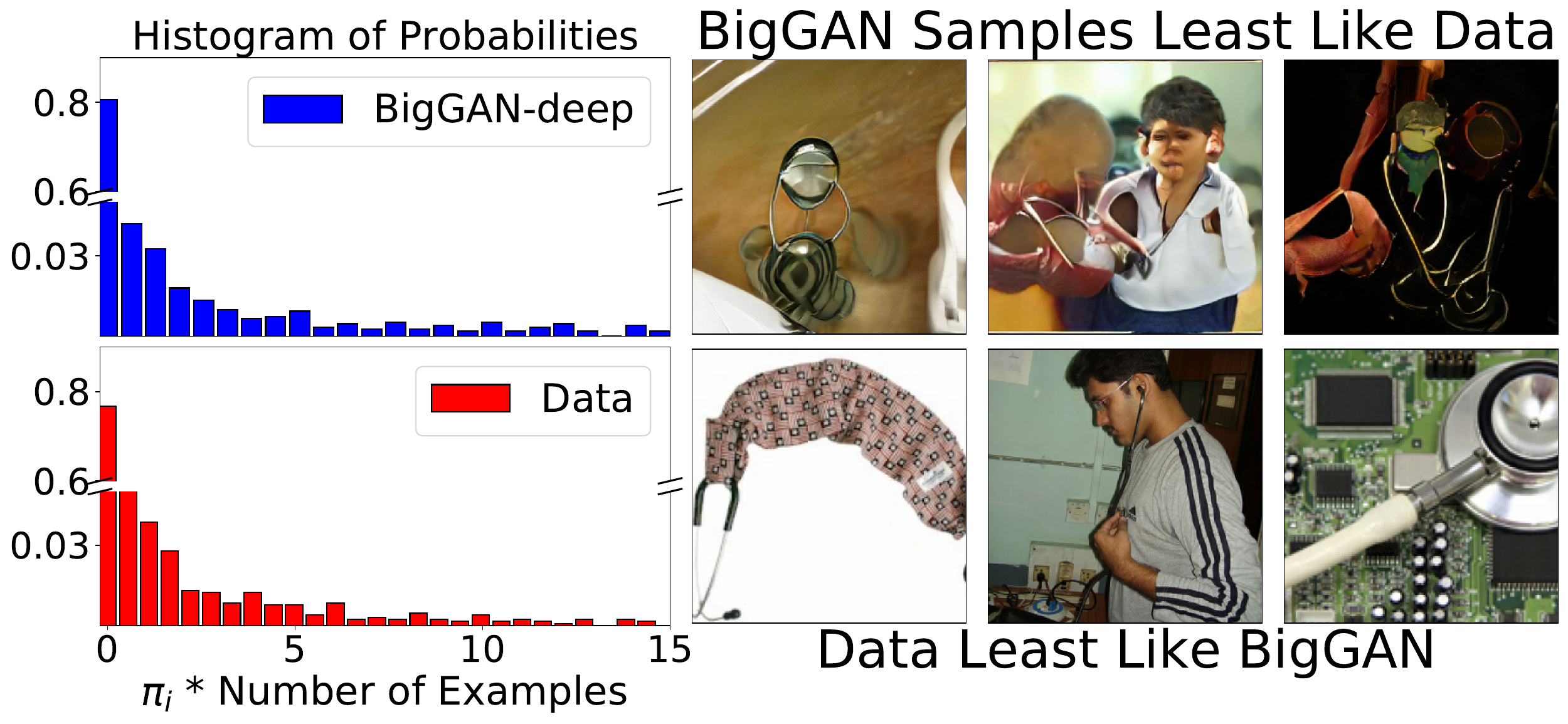} \\
    \leastlikelysinglefigure{Frilled Lizard}{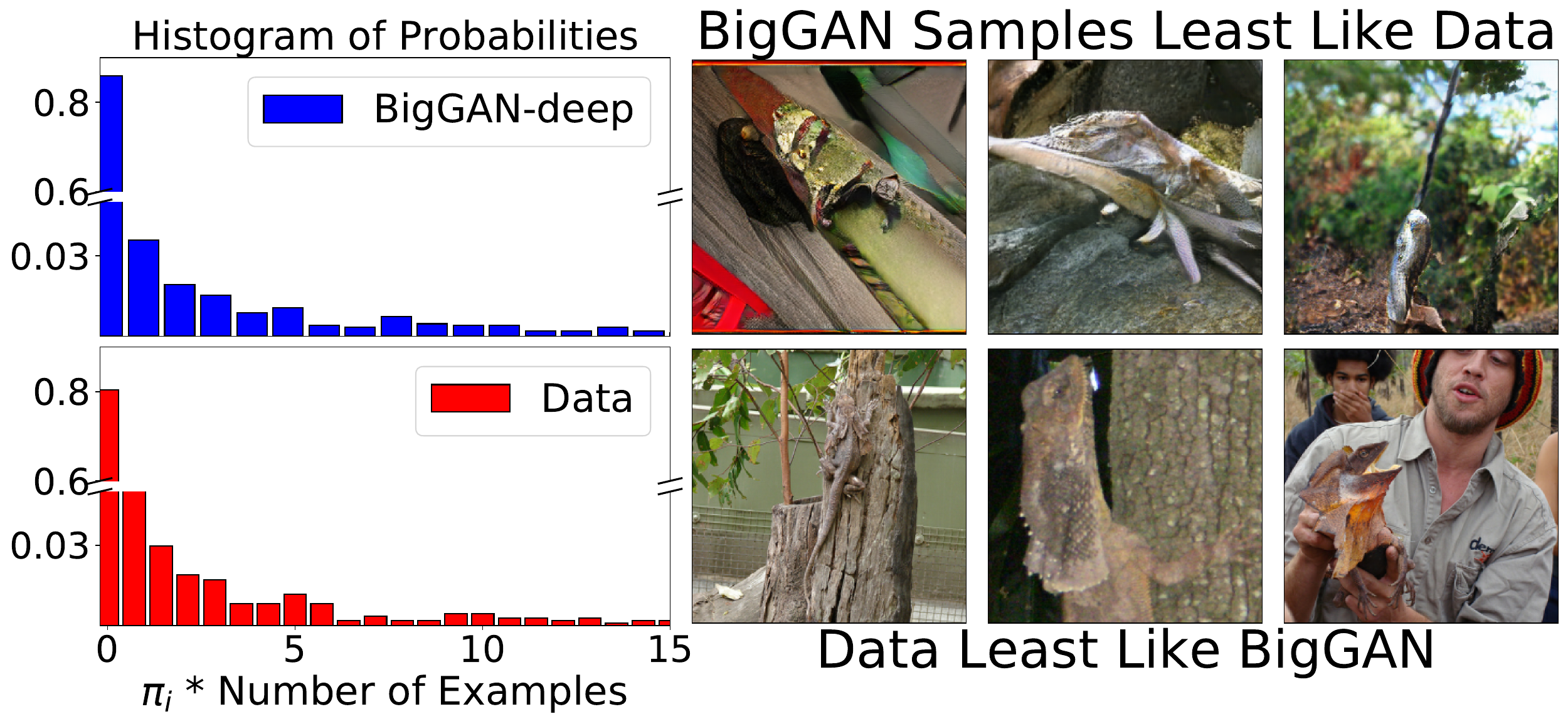} \\
    \leastlikelysinglefigure{Giant Panda}{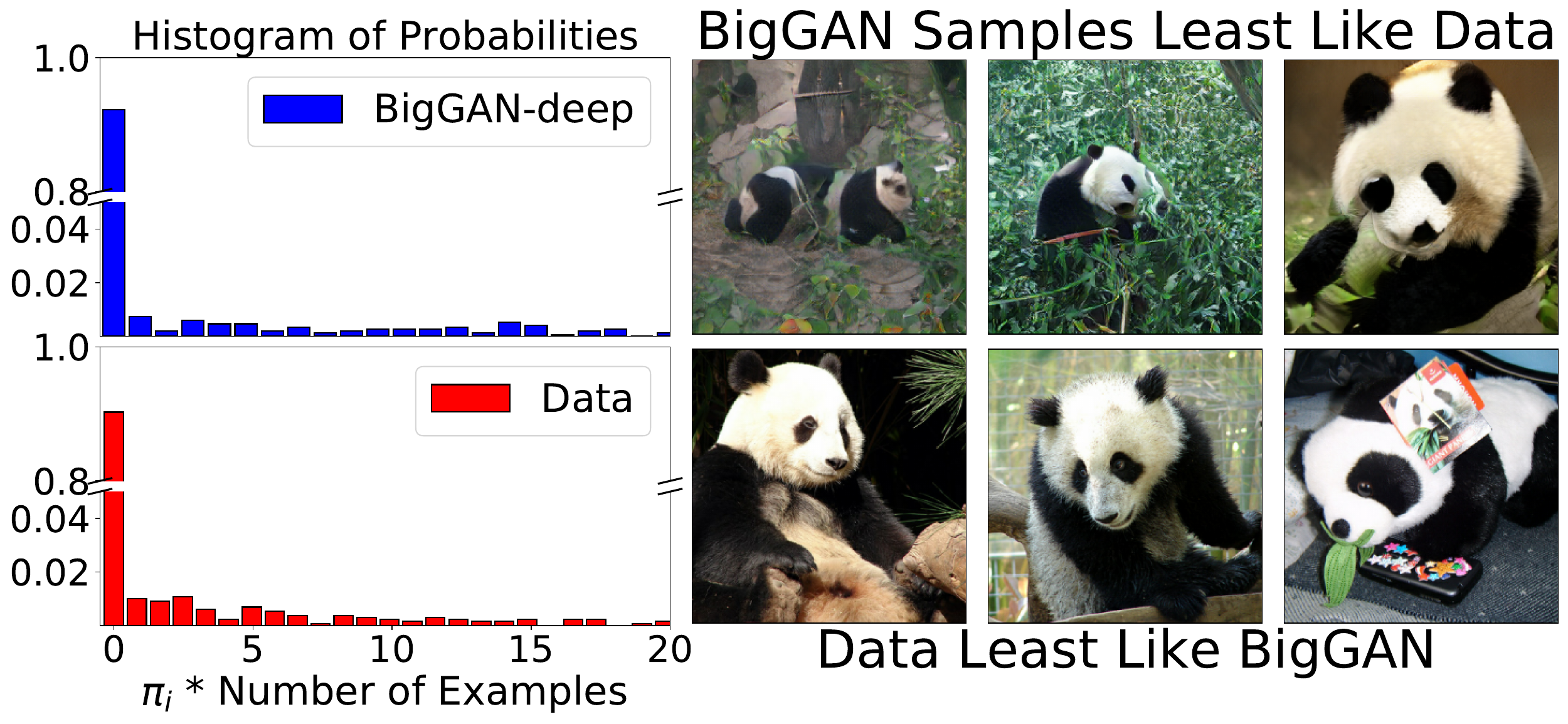} \\
    \end{tabular}
    \setlength{\tabcolsep}{6pt}
    \vspace{-3mm}
    \caption{\suman{Model and test probabilities of the \ktsgel{} test can be used to identify data that the model cannot represent, and samples outside the data distribution. In this example, we look at examples with $0$ model and data probabilities for BigGAN-deep (left) and Cascaded Diffusion Models (right). The blue and red histograms are those for model and data probabilities, respectively. The three top-right images are model samples least like the data distribution ($0$ model probability), and the bottom-right are examples from the data distribution the model cannot represent ($0$ data probability).}}
    \label{fig:app_least_likely_one_figure1}
\end{figure*}
\begin{figure*}[t]
    \centering
    \setlength{\tabcolsep}{0.5pt}
    \begin{tabular}{c c c}
    \leastlikelysinglefigure{Cairn Terrier}{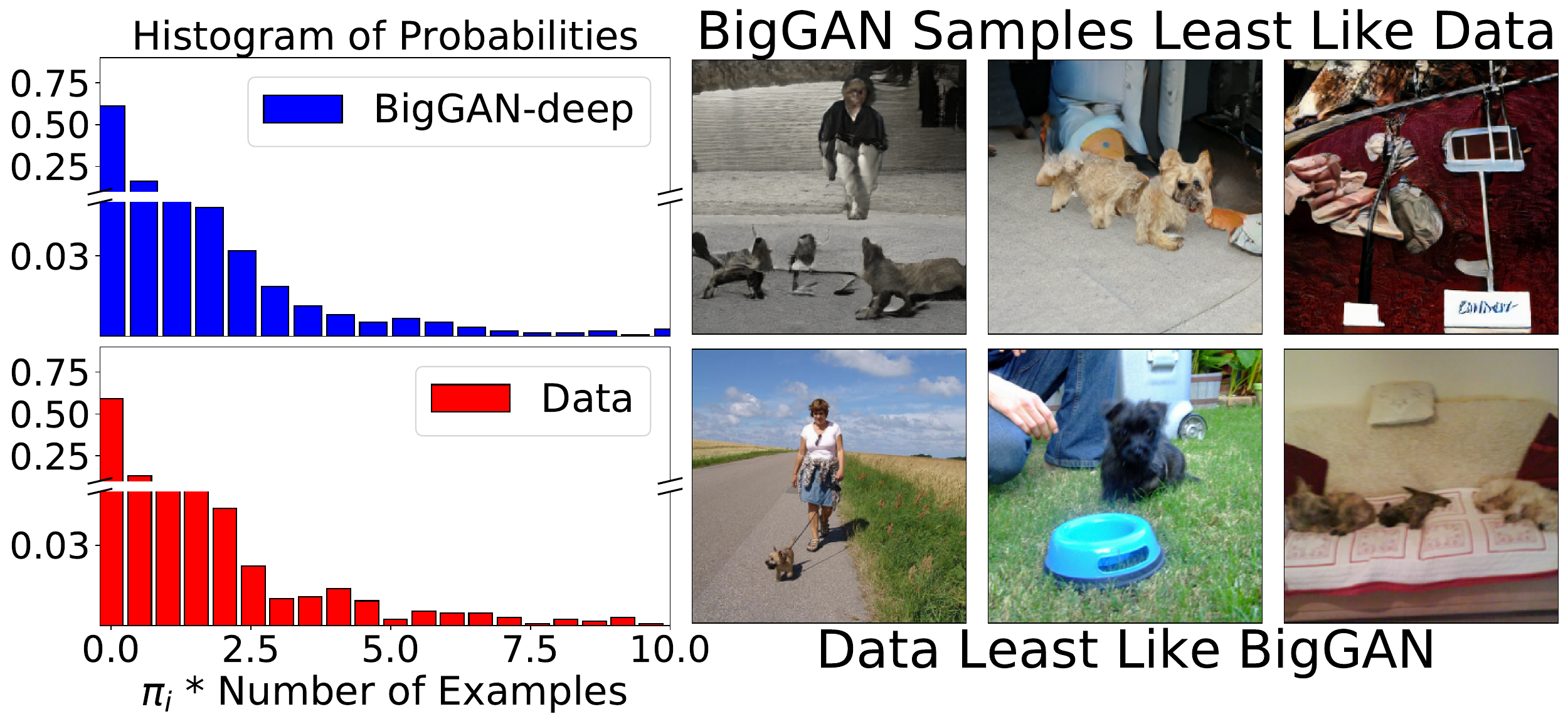} \\
    \leastlikelysinglefigure{Go-Kart}{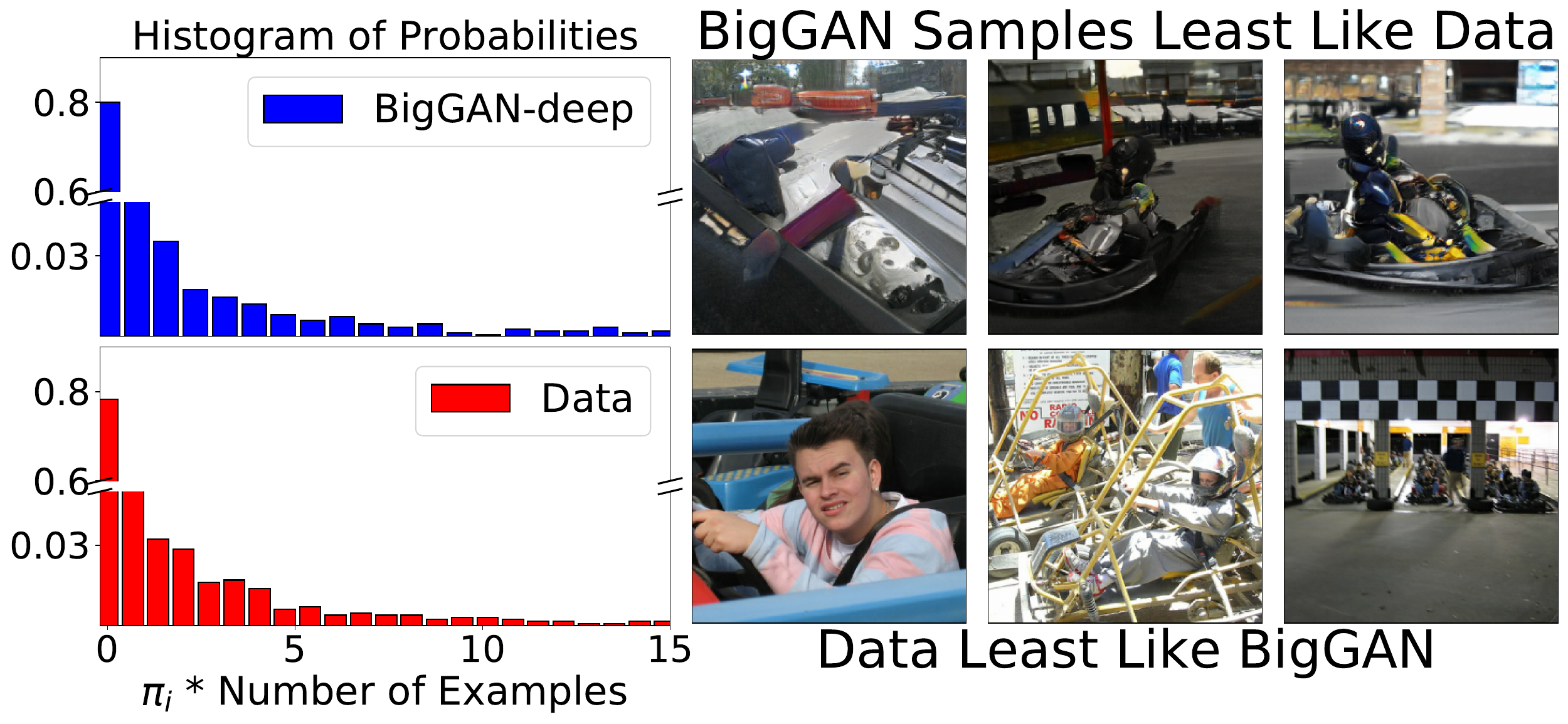} \\
    \leastlikelysinglefigure{Balloon}{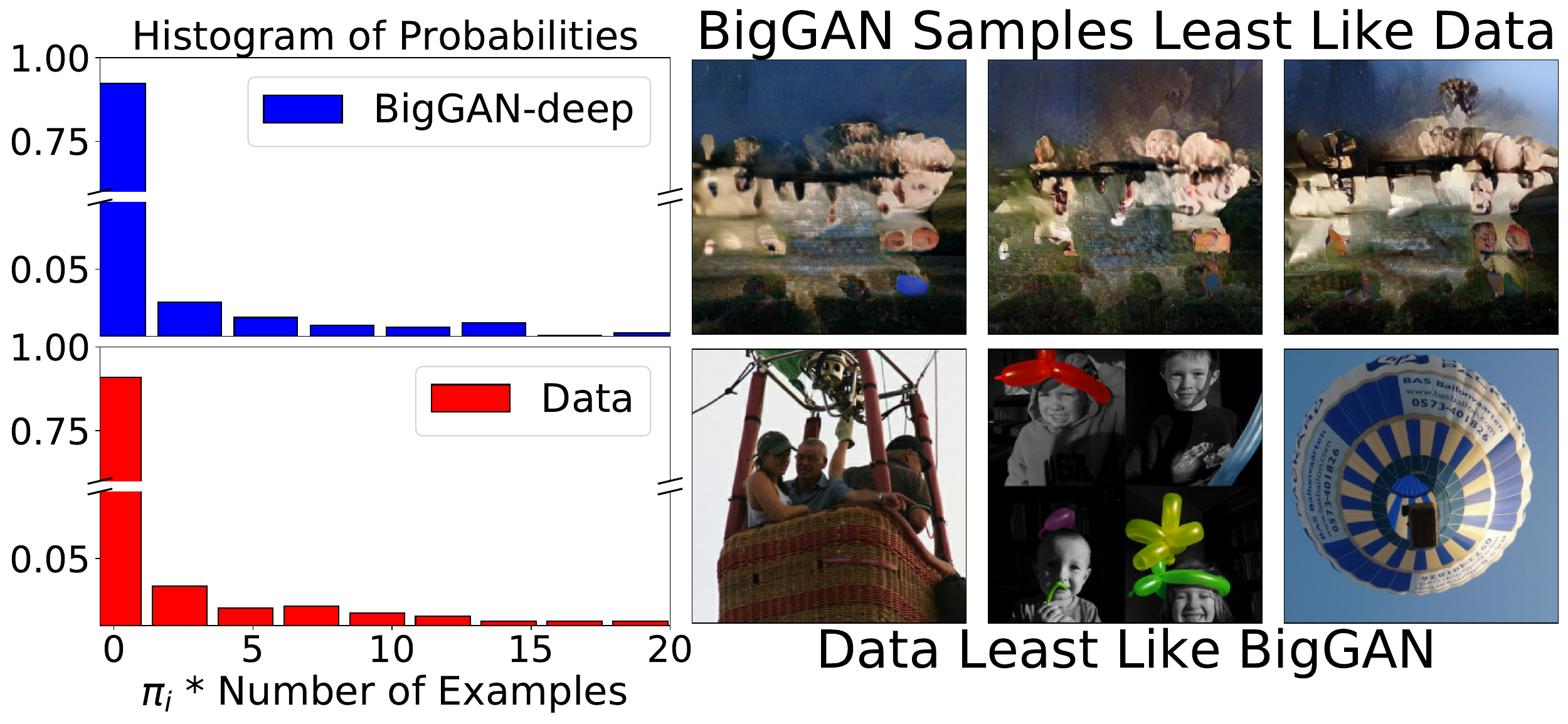} \\
    \leastlikelysinglefigure{Green Mamba}{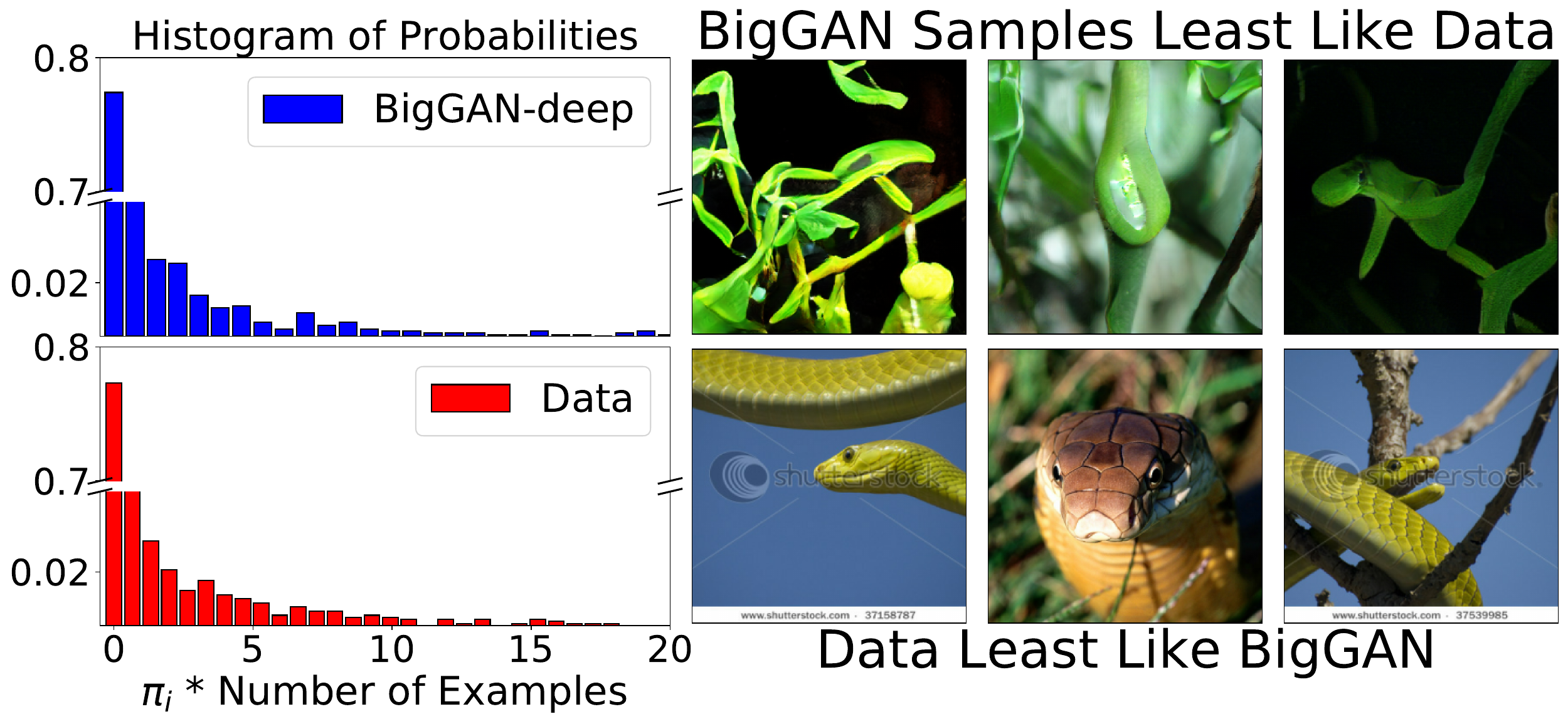} \\
    \leastlikelysinglefigure{Great White Shark}{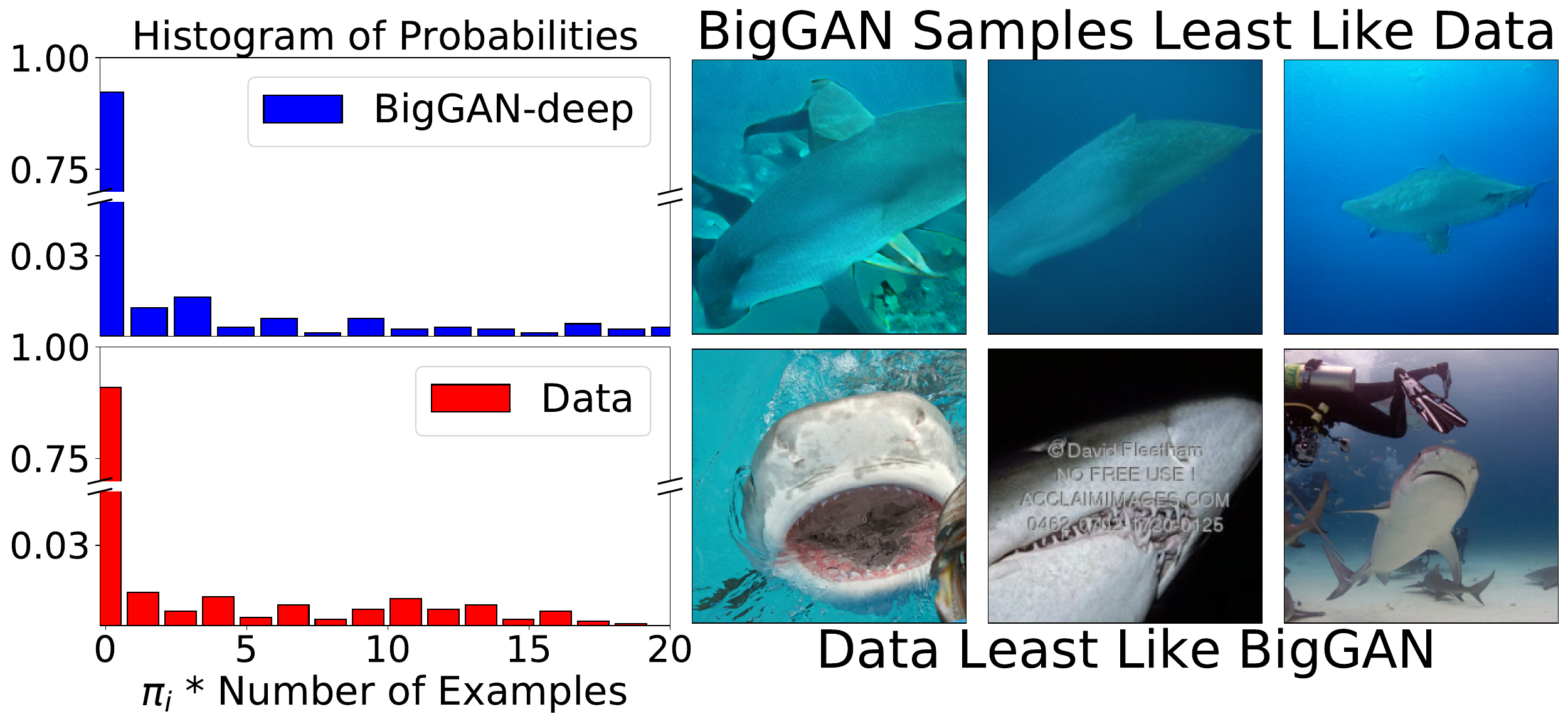} \\
    \end{tabular}
    \setlength{\tabcolsep}{6pt}
    \vspace{-3mm}
    \caption{\suman{Model and test probabilities of the \ktsgel{} test can be used to identify data that the model cannot represent, and samples outside the data distribution. Here, we look at examples with $0$ model and data probabilities for BigGAN-deep (left) and Cascaded Diffusion Models (right). The blue and red histograms are those for model and data probabilities, respectively. 
    The three top-right images are model samples least like the data ($0$ model probability), and the bottom-right are examples from the data distribution the model cannot represent ($0$ data probability).}}
    \label{fig:app_least_likely_one_figure2}
\end{figure*}

%% file: 07_appendix_code.tex
\clearpage
\onecolumn
\section{GEL Iteration Pseudocode}
\subsection{Empirical Likelihood Iteration}
\label{ssec:app_el_iteration}
\begin{minted}{python}
def emp_lik_iteration(feature_diffs, params):
  """Calculate Newton step for empirical likelihood.

  Args:
    feature_diffs: array of m(x_i; c).
    params: current parameters.
  Returns:
    params: parameters after a Newton step.
    probs: updated per-sample probabilities.
    log_lik: updated log likelihood.
    best_log_lik: theoretically optimal log likelihood.
    n_out_of_domain: number of points <0.0 or >1.0.
    log_grad_norm: norm of the gradient (to check optimization).
  """
  num_examples = feature_diffs.shape[0]
  z = 1.0 + np.dot(feature_diffs, params)
  inv_n = 1.0 / num_examples

  # positive part of the modified logarithm
  w_pos = 1.0 / z[z >= inv_n]
  f_diff_pos = feature_diffs[z >= inv_n, :] * w_pos[:, np.newaxis]

  w_neg = (2.0 - num_examples * z[z < inv_n]) * num_examples
  f_diff_neg = feature_diffs[z < inv_n, :]
  num_egs2 = num_examples ** 2

  neg_hess = (np.dot(f_diff_pos.T, f_diff_pos)
              + np.dot(f_diff_neg.T, f_diff_neg) * num_egs2)
  sc_f_diff_neg = f_diff_neg * w_neg[:, np.newaxis]
  log_grad = np.sum(f_diff_pos, axis=0) + np.sum(sc_f_diff_neg, axis=0)
  log_grad_norm = np.linalg.norm(log_grad)

  direction = np.linalg.solve(neg_hess, log_grad)
  params += 1.0 * direction
  n_out_of_domain = f_diff_neg.shape[0]

  probs = 1.0 / (num_examples * (1.0 + np.dot(feature_diffs, params)))

  log_lik = np.sum(np.log2(probs))
  best_log_lik = -num_examples*np.log2(num_examples)
  return params, probs, log_lik, best_log_lik, n_out_of_domain, log_grad_norm
\end{minted}
\newpage
\subsection{Exponential Tilting Iteration}
\label{ssec:app_et_iteration}
\begin{minted}{python}
def exp_tilt_iteration(feature_diffs, params):
  """Calculate Newton step for exponential tilting.

  Args:
    feature_diffs: array of m(x_i; c).
    params: current parameters.
  Returns:
    params: parameters after a Newton step.
    probs: updated per-sample probabilities.
    ent: updated entropy.
    best_ent: theoretically optimal entropy in bits.
    n_out_of_domain: number of points <0.0 or >1.0.
    log_grad_norm: norm of the gradient (to check optimization).
  """
  num_examples = feature_diffs.shape[0]
  w_exp_tilt = np.exp(np.dot(feature_diffs, params)) / num_examples
  sc_f_diff = feature_diffs * w_exp_tilt[:, np.newaxis]

  hess = np.dot(sc_f_diff.T, feature_diffs)
  log_grad = np.sum(sc_f_diff, axis=0)
  newton_step = np.linalg.solve(hess, log_grad)
  log_grad_norm = np.linalg.norm(log_grad)

  params -= 0.5 * newton_step
  exp_weights = np.exp(np.dot(feature_diffs, params))
  probs = exp_weights / np.sum(exp_weights)
  n_out_of_domain = 0
  ent = entropy(probs, base=2)
  best_ent = np.log2(num_examples)
  return params, probs, ent, best_ent, n_out_of_domain, log_grad_norm
\end{minted}